\documentclass[lettersize,journal]{IEEEtran}

\usepackage{cite} 
\usepackage[pdftex]{graphics} 
\usepackage{epsfig} 
\usepackage[cmex10]{amsmath} 
\usepackage{newtxmath} 
\usepackage{times} 
\usepackage[hidelinks]{hyperref} 
\usepackage[all]{hypcap} 
\usepackage[capitalise,nameinlink]{cleveref} 
\usepackage{algorithm}
\usepackage{algpseudocode}
\usepackage{bm}
\usepackage{multirow}
\usepackage[subrefformat=parens,labelformat=parens,caption=false,font=footnotesize]{subfig}
\usepackage{array}
\usepackage{pifont}
\usepackage{balance}
\usepackage[flushleft]{threeparttable}
\usepackage{booktabs}
\usepackage{todonotes}

\newcommand\norm[1]{\left\lVert#1\right\rVert}

\title{
Dexterous In-Hand Manipulation of Slender Cylindrical Objects through Deep Reinforcement Learning with Tactile Sensing
}

\author{Wenbin Hu$^{1}$, Bidan Huang$^{2 \dag}$, Wang Wei Lee$^{2}$, Sicheng Yang$^{2}$, Yu Zheng$^{2}$, Zhibin Li$^{3}$
\thanks{$\dag$ Corresponding author.}%
\thanks{$^{1}$Wenbin Hu is with the School of Informatics, the University of Edinburgh, EH8 9YL, Edinburgh, U.K. This study was conducted during his internship at Tencent Robotics X. (e-mail: wenbin.hu@ed.ac.uk)}
\thanks{$^{2}$Bidan Huang$^\dag$, Wang Wei Lee, Sicheng Yang, Yu Zheng are with Tencent Robotics X, Shenzhen, China. (e-mail: \{bidanhuang, wwlee, sichengyang, petezheng\}@tencent.com)}%
\thanks{$^{3}$Zhibin Li is with the Department of Computer Science, University College London, WC1E 6BT, London, U.K. (e-mail: alex.li@ucl.ac.uk)}
}

\begin{document}

\maketitle

\begin{abstract}
Continuous in-hand manipulation is an important physical interaction skill, where tactile sensing provides indispensable contact information to enable dexterous manipulation of small objects.
This work proposed a framework for end-to-end policy learning with tactile feedback and sim-to-real transfer, which achieved fine in-hand manipulation that controls the pose of a thin cylindrical object, such as a long stick, to track various continuous trajectories –- through multiple contacts of three fingertips of a dexterous robot hand with tactile sensor arrays. 
We estimated the central contact position between the stick and each fingertip from the high-dimensional tactile information and showed that the learned policies achieved effective manipulation performance with the processed tactile feedback.
The policies were trained with deep reinforcement learning in simulation and successfully transferred to real-world experiments, using coordinated model calibration and domain randomization.
We evaluated the effectiveness of tactile information via comparative studies and validated the sim-to-real performance through real-world experiments.
\end{abstract}
\section{Introduction}\label{sec:introduction}

\begin{figure}
    \centering
    \includegraphics[trim=150 50 200 0, clip, width=\linewidth]{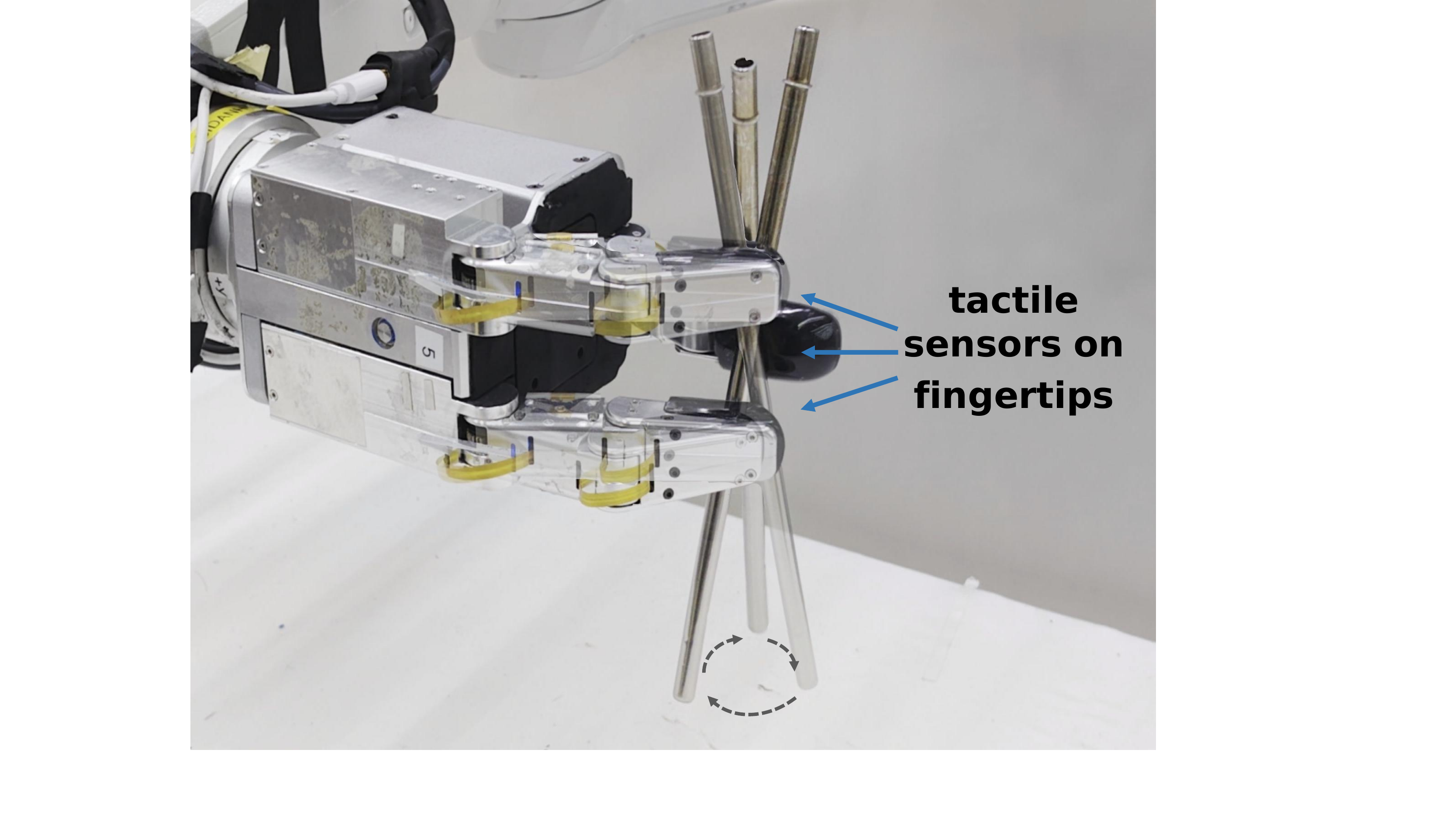}
    \vspace{-5mm}
    \caption{Steering a stick with a robot hand equipped with tactile sensor arrays on fingertips. The control policy is trained in simulation using deep reinforcement learning and directly transferred to the real robot.}
    \label{fig:first_page}
    \vspace{-4mm}
\end{figure}

The dexterous manipulation of in-hand objects is a skill that humans possess effortlessly. Such manipulation involves moving only our fingers, without excessive wrist or arm motion, to alter the pose of objects. The ability to endow robots with such athletic intelligence is highly desirable, particularly when working in a confined space or using certain tools, such as stirring rods.
However, it still remains challenging for robots to maintain and secure a firm grasp while continuously changing the object pose.

Unlike manipulating large objects where the robot fingertips can be approximated as point contacts, manipulation of small objects (e.g. slender cylindrical objects) requires highly coordinated finger motion and precise perception of the contacts.
As the feasible contact areas are small and narrow, in-hand manipulation is more sensitive to errors and uncertainties.
In this paper, we explored the in-hand manipulation ability of a three-fingered robot hand equipped with tactile sensors to address these challenges involved in manipulating cylindrical objects, as shown in \cref{fig:first_page}.

When humans manipulate objects, contact events are captured in exquisite detail by dense mechanoreceptors embedded in the skin, offering vital contact information such as the time, position and forces experienced. In robots, similar information can only be effectively captured by tactile sensors, since other sensors such as vision and proprioception may be occluded by fingers or overwhelmed by background noise \cite{2020_tactile_review}. 

A variety of tactile sensors are available today, such as the vision-based tactile sensors \cite{2022_nathan, 2020_digit, 2019_gelsight} and distributed tactile sensor arrays \cite{2022_uskin}. Real-time spatial and kinetic relations between the hand and the manipulated object can be derived from the signals gathered by tactile sensors, providing a natural way to regulate the manipulation control loop \cite{2021_queen}.

In this paper, we analyzed different ways of exploiting the signals from distributed tactile sensors. We used the estimated positions of contact centers as tactile feedback instead of the raw data from the tactile sensors, because the low-dimensional tactile information alleviates the difficulties of sensor modeling and feature extraction.
The comparative study showed that the policy trained with central contact positions outperforms baselines using other perception of the manipulated object.

Recently, many methods have been proposed to perform in-hand manipulation tasks. Model-based trajectory optimization achieved good performances on object reorientation tasks with both under-actuated hands \cite{2017_vision_mpc, 2016_adaptive_hand} and fully-actuated hands \cite{2018_kenematic_TO, 2014_todorov}. However, the high-dimensional search space presented by multi-fingered robot hands results in optimization problems that are difficult to be solved in real time. The errors and uncertainties of the hand dynamics and contact model also limit the planning performance in real-world experiments.

Alternatively, model-free deep reinforcement learning (DRL) have been successfully implemented on various grasping and manipulation tasks \cite{2022_finger_gait, 2022_torque_controlled_hand, 2020_openai}, where the policies are learned from scratch in a trial-and-error manner. 
In this work, we utilized model-free DRL to train dexterous tactile-based manipulation skills.
Policies trained with pure simulated data often suffer the performance decline when transferred to real-world experiments, and therefore we implemented model calibration of the finger joints and randomized the initial states to mitigate the sim-to-real gap.

During the manipulation of a slender cylindrical object, it is crucial for all three fingertips to remain in contact with it to avoid dropping. Therefore the finger-gaiting solution where the fingers are making and breaking contacts with the object by turns \cite{2022_torque_controlled_hand} is not viable here. Instead, our robot manipulated the stick by continuously changing the contact locations, and sliding it across the curved fingertip surfaces without breaking any contact.
Our proposed method is validated through the continuous stick pose tracking tasks, as shown in \cref{fig:first_page}.

The contributions of this work include: 
\begin{itemize}
\item[(1)] A proposed sim-to-real transfer method that accounts for mechanical backlash, characterizes finger joint dynamics, and uses domain randomization for initial training states to reduce the sim-to-real gap and enable successful real-world deployment of learning-based policies;
\item[(2)] Effective modeling and data processing for tactile sensors in physics simulation and real tactile feedback to enable dexterous manipulation of slender cylindrical objects, demonstrating the effectiveness of tactile sensing through a comparative study;
\item[(3)] Autonomous, continuous in-hand manipulation of slender cylindrical objects with controlled pose tracking using only 3 fingertips with tactile sensor arrays on a robot hand, demonstrating the potential for robotic dexterity in constrained spaces.
\end{itemize}

This paper is organized as follows. In \cref{sec:related_work} we reviewed the related work. The real and simulated robot hand with tactile sensors were described in \cref{sec:hardware}. We introduced the details of policy learning and sim-to-real techniques in \cref{sec:policy_learning} and \cref{sec:sim2real_transfer}. The evaluation of the learned policies in both simulation and real-world experiments were presented in \cref{sec:experiments}. We concluded the paper and proposed future research directions in \cref{sec:discussion}.
\section{Related Work}\label{sec:related_work}

\textbf{Tactile-based manipulation.} Tactile provides indispensable and the most direct sensory information of the contacts in dexterous grasping and manipulation tasks. 
For learning-based algorithms, though the joint position and torque provide feasible feedback for adaptive grasping \cite{2022_wang}, tactile information improves both the performances and the sample-efficiency of policy training \cite{2021_sample_efficiency}.
Vision-based tactile sensors like GelSight \cite{2017_gelsight} and DIGIT \cite{2020_digit} render the deformation of the hand surfaces as high resolution images, which can be used to extract feature vectors \cite{2018_more_than_feeling, 2020_digit}, generating grasping pose \cite{2023_jiang}, or predict future frames \cite{2019_gelsight}.
Simulation of such tactile sensors and the sim-to-real transfer is challenging \cite{2022_tactile_gym, 2022_tacto}.
In this paper we used the distributed tactile sensor arrays, which are easy to deploy on robot hands and can cover more potential contact areas. Graph convolutional network (GCN) is feasible for combining the irregularly aligned tactile data on different parts of the hand together while maintaining the positional relation of the sensors \cite{2022_uskin}.

During the in-hand manipulation of the cylindrical objects, only a small fraction of the distributed tactile sensors are in contact simultaneously, because the stick is thin and the inner surfaces of the fingertips are curved. Hence, the raw tactile data is sparse, and directly using such tactile data for end-to-end policy learning requires a large training data set or a complicated network structure such as GCN.
Instead, the processed information of contact positions is low-dimensional and more intuitive, and it is proved to be effective in grasping \cite{2019_MAT} and in-hand manipulation \cite{2022_finger_gait}. Lambeta et al. trained an auto-encoder to extract the contact positions from tactile images \cite{2020_digit} and Wang et al. estimated the contact positions from distributed tactile arrays \cite{2019_wang}.
In this work, similar to \cite{2019_wang}, we computed the position of the contact center at each fingertip as the tactile feedback.

\textbf{Deep reinforcement learning} has been successfully implemented in dexterous in-hand manipulation \cite{2020_openai}. Filipe et al. proposed a hierarchical structure where a high-level DRL-based controller is combined with a low-level grip stabilizer \cite{2020_hierarchical}. Manipulation skills using only the fingertips can also be learned with DRL, e.g. finger-gaiting and finger-pivoting \cite{2022_torque_controlled_hand, 2022_finger_gait}.
In this work, we used DRL to learn a more challenging task: continuous fingertip manipulation of the slender cylindrical objects where the contact areas are narrow and can be moving on the finger surfaces.

\textbf{Sim-to-real transfer}.
Leveraging the physics simulator, the learning process can be dramatically accelerated. However, the discrepancies between the simulated and real environments are the major cause of the performance declines when the convergent policies transferred to reality.
System identification (SI) mitigates the sim-to-real gap by modeling the system dynamics and calibrating the parameters by experiments \cite{2018_nonprehensile, 2022_torque_controlled_hand}.
Domain randomization (DR) trades the policy optimality for generalization ability, by randomizing the dynamics of simulated environment and perception space \cite{2021_trifinger, 2020_openai, 2021_droid}. 
In this work, we utilized SI to calibrate the dynamic features of finger joints, and DR to randomize the physical properties that are difficult to model, e.g. friction coefficients between target object and fingers. We also randomized the initial states of the manipulated object and fingers at the beginning of each training episode, to increase the ability of the policies to handle disturbances and uncertainties.
\section{Three Finger Robot Hand with Tactile Sensors and Simulated Models} \label{sec:hardware}
\begin{figure}[t!]
    \centering
    \includegraphics[trim=100 85 100 90, clip, width=\linewidth]{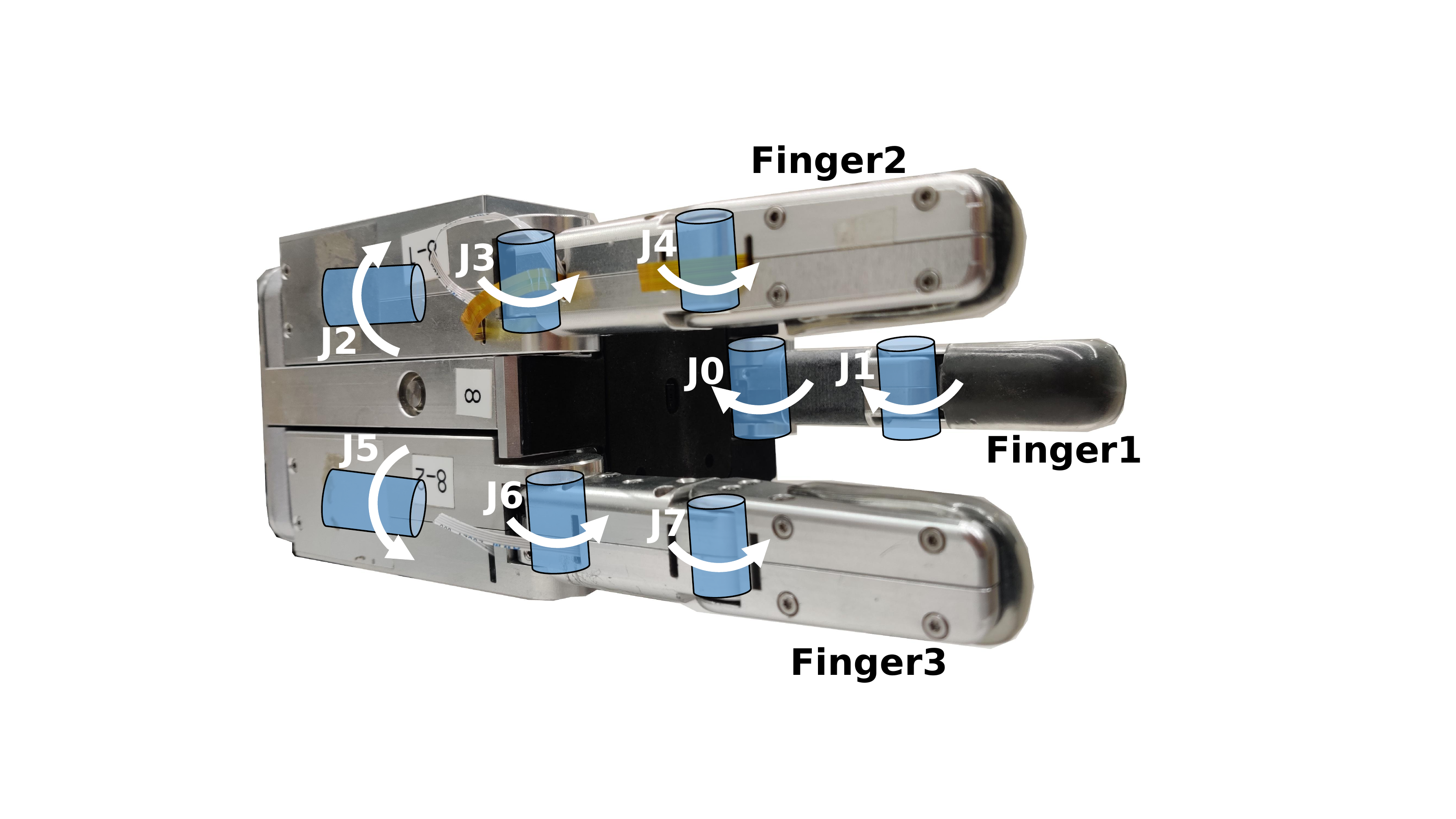}
    \caption{Kinematic diagram of the TRX Hand, showing all 8 joints.}
    \label{fig:hand_joints}
    \vspace{-4mm}
\end{figure}
\subsection{Real Robot Hand with Tactile Sensors} \label{subsec:real_hand}
This work used a custom build three-fingered robot hand. Named the TRX hand, it comprises of 8 fully-actuated joints, as shown in \cref{fig:hand_joints}. Each finger has two joints, while the base of fingers 1 and 2 are mounted on additional rotation joints (J2 and J5) that allow them to spread independently about the palm up to 90 degrees. 
J0, J3 and J6 are transmitted by the worm gear mechanism and therefore have backlash and self-lock features. We simulated these features as described in \cref{sec:sim2real_transfer:joint}.
The maximum resultant fingertip force is 15N for each finger.
Unlike many other robot hand designs using flat fingertips, the inner surfaces of TRX fingertips are deformable and curved, which are more human-like and conducive to dexterous in-hand manipulation.

All three fingertips are equipped with tactile sensors. These sensors are also developed in-house and feature an array of piezoresistive sensing elements (taxels) distributed across a continuous, curved surface. The sensor arrays are covered by an additional layer of silicone material to provide about 1\,mm of mechanical compliance. There are 128 taxels per fingertip. Each tactile taxel returns a value that is proportional to the applied normal force on it.

\subsection{Simulated Robot Hand and Tactile Sensors} \label{subsec:simulated_hand}
\begin{figure}[t!]
    \centering
    \includegraphics[trim=270 0 170 0, clip, width=0.7\linewidth]{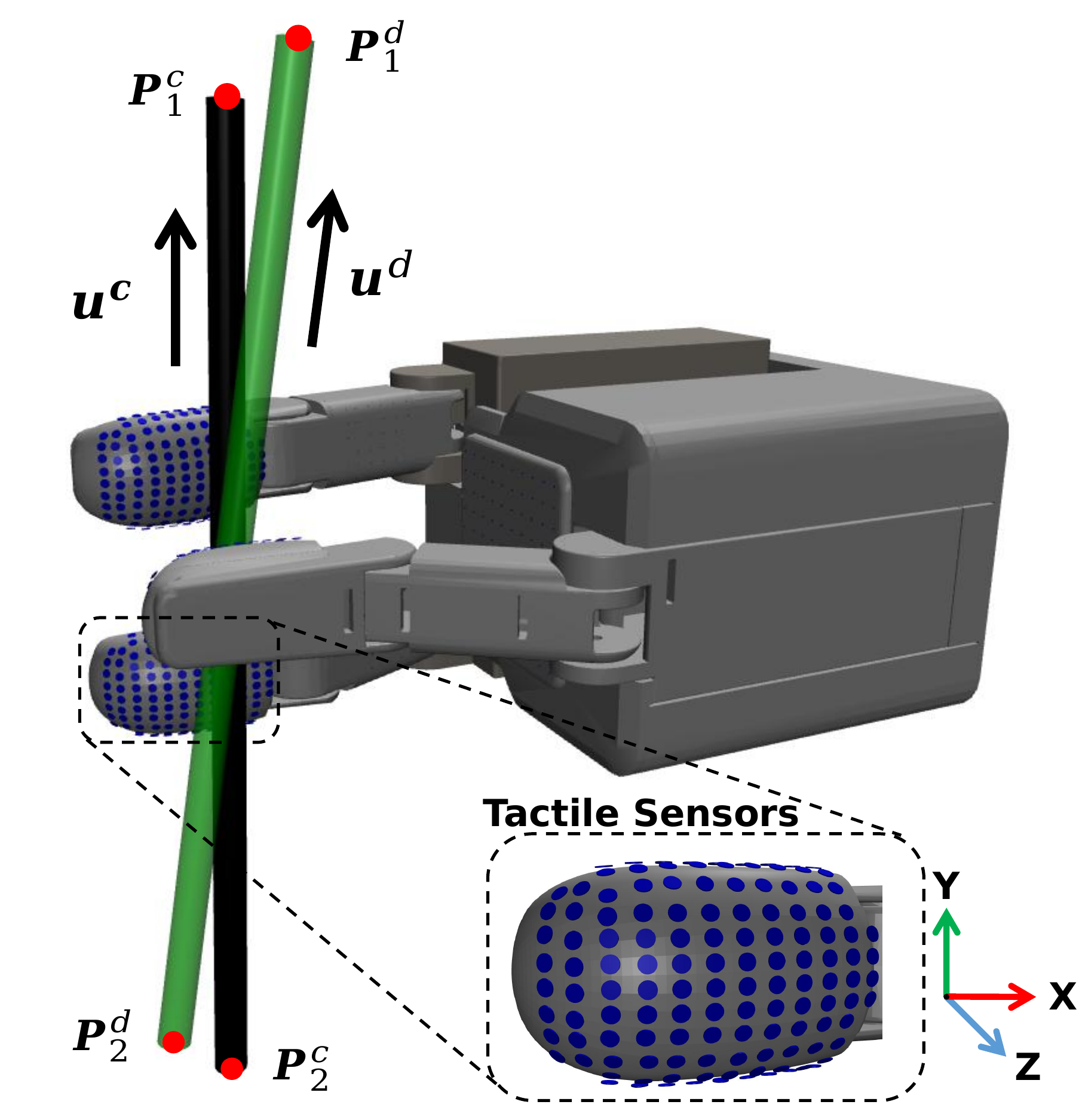}
    \caption{The real pose (black) and the desired pose (green) of the target stick. The learned policy aims to manipulate the stick to track the desired continuous pose in real time. $\mathbf{u}$ denotes the 3-dimensional unit vector of the major axis, and $\mathbf{P}_i$ denotes the key-point of the stick.}
    \vspace{-4mm}
    \label{fig:reference_pose}
\end{figure}

We used MuJoCo physics engine \cite{MuJoCo} to train the manipulation policies which has fast, accurate simulation and tunable soft contact constraints.
\cref{fig:reference_pose} shows the environment with the robot hand and the target.
The wrist of the hand is fixed at a certain pose in the mid-air throughout the manipulation. 
Though the TRX hand has 8 joints, in this paper we only used 6 of them, fixing J2 and J5 at zero poses, because empirically the human index and middle fingers do not rotate laterally when manipulating a stick.
At the beginning of each episode, the finger joints are set to initial positions and the object is placed at the initial pose and the gravity applied on the object is compensated for a short time (0.1s), allowing the fingers to close and establish contacts.

The virtual distributed tactile sensors were simulated as the small, thin, and light cylinders located at the finger surfaces shown as the blue dots in \cref{fig:reference_pose}, similar to the sensor modeling used in \cite{ding2021sim}. 
Replicating the sensor features in simulation minimizes the sim-to-real gap, and exploits the full potential of tactile sensors. 
However, in the manipulation tasks where the stick is rigid, the measurement of the exact magnitude of normal forces is less critical.
Meanwhile, modeling the non-linear relationship between the tactile signal and normal force applied to the sensor is non-trivial in simulation, and including such training the end-to-end policies require a large amount of real tactile data.

Therefore, in simulation, each sensor returns the binary signal of whether it is in contact with the target object, while in real experiments we binarized the tactile signals to Boolean values. Despite MuJoCo only supports point rigid body contact, with the attached sensors we are able to approximate the line or plane contact with multiple contact points.

\section{Policy Learning in Simulation} \label{sec:policy_learning}

\subsection{Task specification}
The task is to manipulate the slender sticks with three fingers to track the real-time target pose trajectories, as shown in \cref{fig:reference_pose}, where the black stick should track the green reference pose. The manipulation only involves the finger motion, and the robot hand is fixed at a certain pose where the palm is perpendicular with the ground during the manipulation. We chose four different stick trajectories as the references, namely using the end of the stick to draw a line, a circle, a spiral and the number of 8, and trained four corresponding policies to track them.

\subsection{Deep Reinforcement Learning}
We modeled the manipulation task as a finite-horizon discounted Markov decision process and used deep reinforcement learning algorithm to learn the control policy. The process consists of an action space $\mathcal{A}$, a state space $\mathcal{S}$, the state transition dynamics $\mathcal{T}:\mathcal{S}\times\mathcal{A}\rightarrow{}\mathcal{S}$ modeled by the physics simulator, and a reward function $r: \mathcal{S}\times\mathcal{A}\rightarrow{}\mathbb{R}$.
At every time-step, the policy $\pi$ gets an observation of the current state $s_t$ and generates the action $a_t=\pi(s_t)$. After the robot interacts with the environment, it receives a reward $r(s_t, a_t)$. The target of the policy is to maximize the expected discounted sum of rewards $\varmathbb{E}_\pi\left[\sum_{t=0}^{T-1}\gamma ^tr(s_t,a_t)\right]$, where $\gamma$ is the discounting factor. We chose Proximal Policy Optimization (PPO) as the DRL algorithm as it is stable and can be easily parallelized.

\subsection{Observation and action space}
The observation space includes three parts: (a) the measured finger joint positions, (b) a unit vector $\mathbf{u}^d=\frac{\mathbf{P}^d_1-\mathbf{P}^d_2}{\norm{\mathbf{P}^d_1-\mathbf{P}^d_2}}$ expressing the major axis direction of the desired stick pose, where the $\mathbf{P}^d_i$ denotes the stick end-point as shown in \cref{fig:reference_pose}, and (c) the position of contact center on each finger, which is computed as the mean position of all the tactile sensors in contact.
The finger joint positions are normalized to $[0, 1]$ by the lower and upper limits.
Since we ignore the stick rotation around its major axis, the unit vector is adequate in representing the stick orientation.
The positions of tactile sensors are in the local coordinate of the corresponding finger link, and normalized by a constant scale factor.
The output actions are the desired displacements of finger joints.

\begin{figure}[tb]
    \centering
    \includegraphics[trim=0 40 0 35, clip, width=\linewidth]{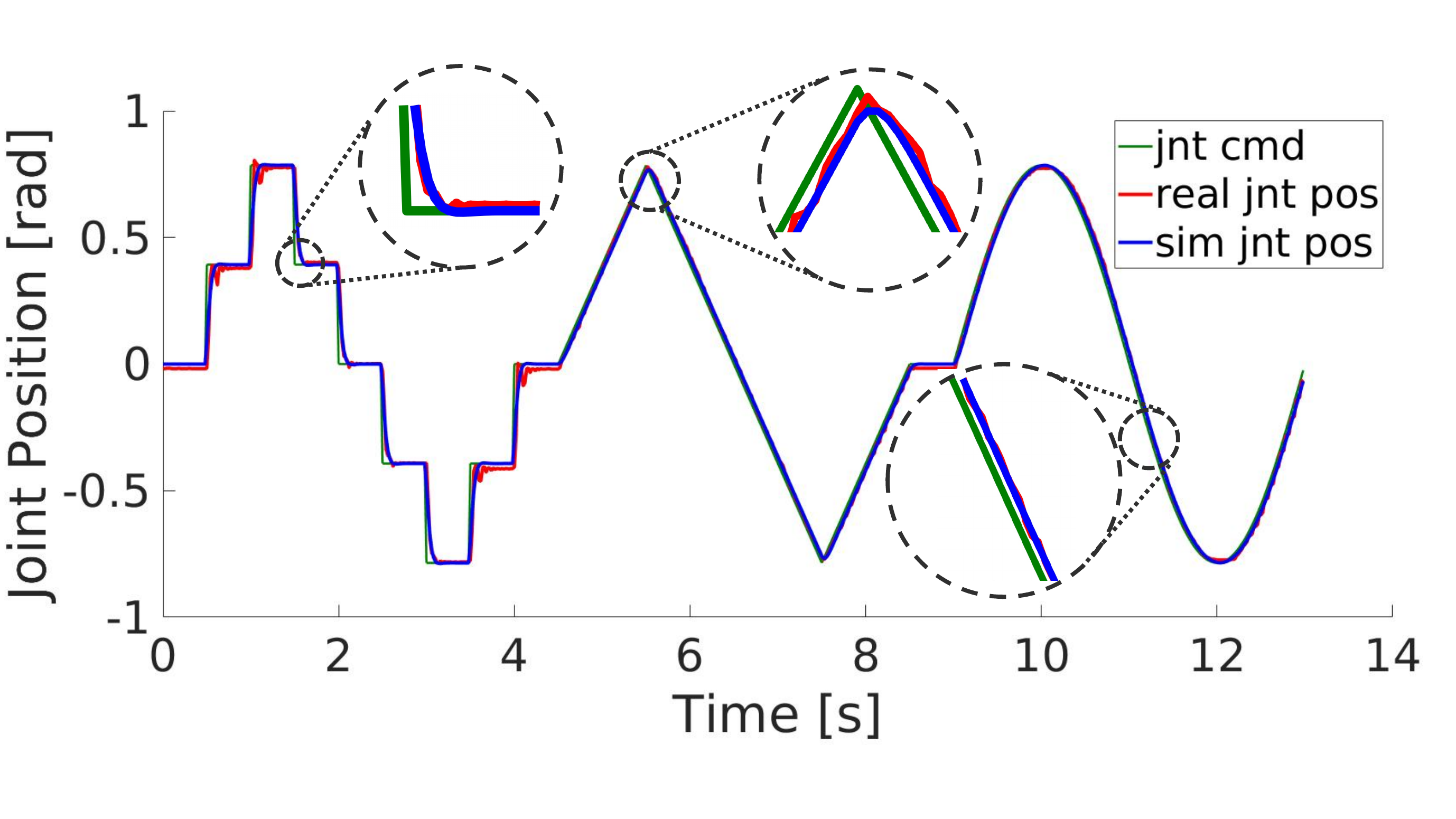}
    \caption{Trajectory-tracking performance with reference and measured joint positions in both simulated and real experiments.}
    \label{fig:cmaes_jnt_traj}
    \vspace{-4mm}
\end{figure}

\subsection{Reward design}
To alleviate the excessive dependence on human prior knowledge, the reward function design is straight-forward, formulated as the weighted sum of four terms:
\begin{equation}
    R = C - \omega_0\norm{\mathbf{u}^d - \mathbf{u}^c}_2 - \omega_1\sum_{j=1}^2{\norm{\mathbf{P}^d_j-\mathbf{P}^c_j}_2}-\omega_2\sum_{i=1}^N{f_i}
\end{equation}
where $C$ is a positive task reward. $\mathbf{u} \in \mathbb{R}^3$ is a unit vector indicating the stick's major axis. $\mathbf{P}_1,\mathbf{P}_2 \in \mathbb{R}^3$ are the positions of stick key-points, namely the two endpoints (see \cref{fig:reference_pose}). Superscripts $d$ and $c$ denote the \textit{desired} and \textit{current} values, respectively. $f_i$ is the magnitude of contact force on each tactile sensor, and
$\omega_0, \omega_1, \omega_2$ are the weights. We terminated the training episode if the height of the stick's geometry centre was below a threshold, indicating that it fell out of the grasp. Hence the constant term $C$ promotes the learning of stable grasping by rewarding longer episode length. The second term penalizes the orientation error, and the third term penalizes the position error between the desired and measured stick pose, where the errors are formulated as the L2-norm. These two terms guide the policy learning by direct quantification of the pose error.
To avoid the damage on hardware caused by excessive forces, the last term regulates the sum of contact forces detected by the tactile sensors, promoting the gentle manipulation skills.
In this work we used $[0.5, 1.5, 2.0, 0.005]$ for $[C, \omega_0, \omega_1, \omega_2]$.

\section{Sim-to-real Transfer}\label{sec:sim2real_transfer}
For precise in-hand manipulation tasks, the discrepancy between simulation and reality is mainly introduced by inaccuracies in joint modelling, differences in sensory feedback, and errors in physical parameters estimation. 
In this section, we show that our proposed method can mitigate the most difficult sim-to-real gap which is the mechanical backlash, and thus increase the generalization ability of trained policies for the direct sim-to-real transfer.

\begin{figure}
    \centering
    \includegraphics[width=0.7\linewidth]{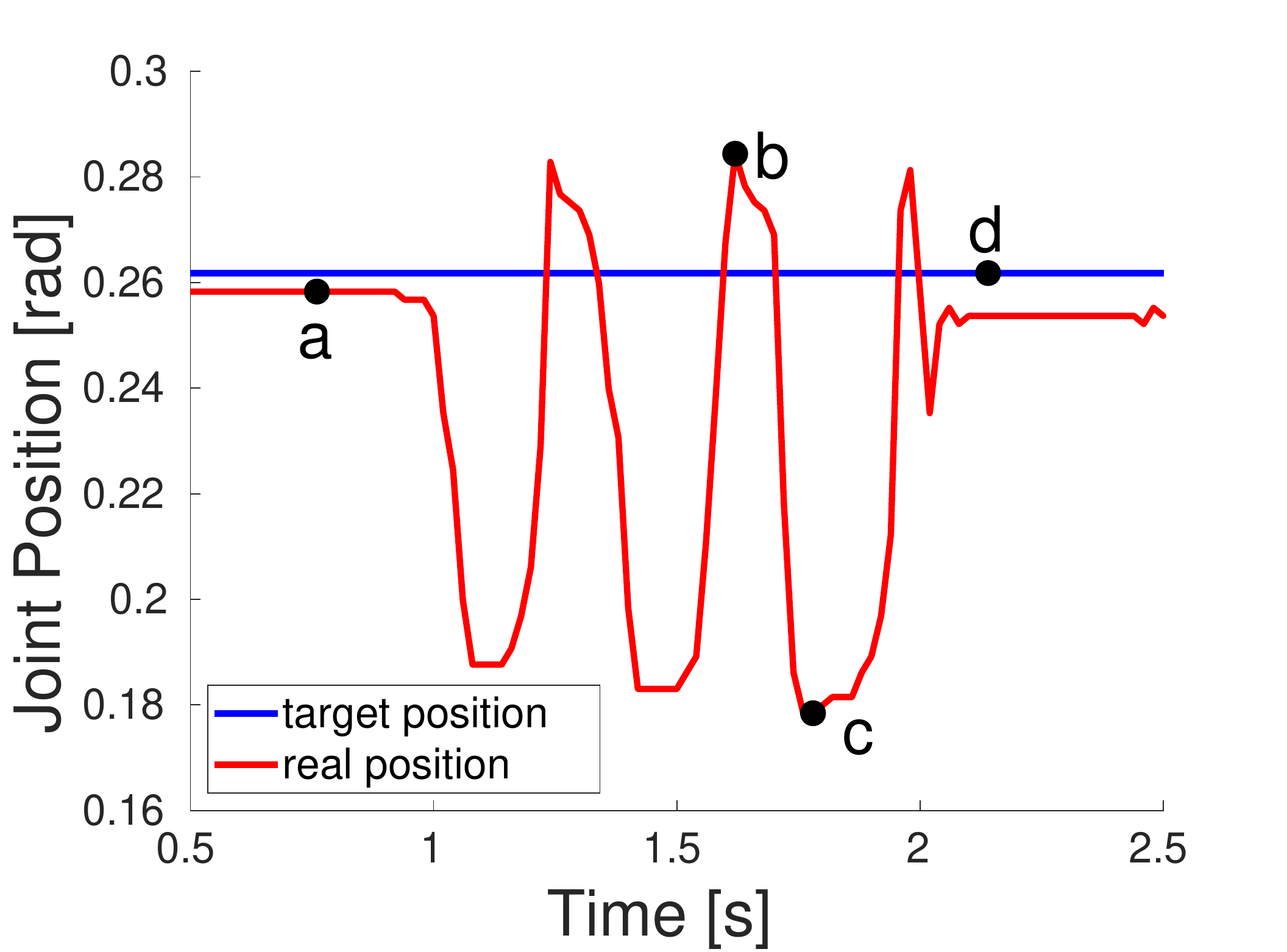}
    \caption{Characterisation and calibration of the mechanical backlash of robot joints. The joint is controlled to maintain a target position and then pushed by external torques. Errors between the static position $q_a$ and extreme positions ($q_b$ and $q_c$) are regarded as the magnitude of the backlash at the target position $q_d$.}
    \label{fig:backlash}
    \vspace{-4mm}
\end{figure}

\subsection{Joint Model Calibration} \label{sec:sim2real_transfer:joint}
The robot hand is position-controlled, thus in simulation the applied torque of each joint is calculated with proportional-derivative (PD) control:
\begin{equation}
    \tau = K_P(q_d-q)-K_D\dot{q}
\label{eq:PD}
\end{equation}
where $K_P$ is the proportional gain and $K_D$ is the derivative gain. $q_d$ denotes the target joint position and $q$ denotes the measured joint position. 
To calibrate the PD gains of each finger joint in simulation, we controlled the target finger joint to track the same reference trajectory in both simulation and real world, and then we compared the recorded joint position trajectories of the simulated robot and real robot, as shown in \cref{fig:cmaes_jnt_traj}. 
We optimized the simulated PD gains by minimizing the errors between the position trajectories of the simulated finger joint and real finger joint:
\begin{equation}
\begin{aligned}
    \min_{K_P, K_D} \quad & \sum_{t=0}^{T}\norm{q^{sim}_t(K_P, K_D)-q^{real}_t}_2 \\
    \textrm{s.t.} \quad & K_P, K_D > 0
\end{aligned}
\end{equation}
where $q^{sim}_t, q^{real}_t$ denote the joint position in both simulation and reality at time-step $t$, and $T$ denotes the trajectory length. We repeated the process for each finger joint to optimize the PD gains respectively.
We utilized the CMA-ES \cite{CMAES} optimization algorithm in this paper.
As shown in \cref{fig:cmaes_jnt_traj}, the calibrated simulated joints have similar dynamics responses to step and continuous signals.

Backlash is the physical gap between mechanical gears that causes an amount of lost motion due to clearance or slackness when gear movement is reversed and contact is re-established. It is an inevitable characteristics for gear driven joints caused by the small gaps between gears and gearbox. With the existence of backlash, at certain positions or during reversing motions, the joint will be out of control within the gaps between gears.
To model such a feature is essential for setting up the realistic simulation and thus learning successful policies, therefore we specifically modeled such uncontrolled backlash joints alongside the actuated joints in simulation, and the actual rotation between the parent and child link is the sum of both joint positions. 

The ranges of the backlash joints were calibrated by the experiments: We controlled the target joint to maintain a certain position $q_d$, and then applied external torques to rotate it as much as possible.
As shown in \cref{fig:backlash}, the static joint position $q_a$ and extreme joint positions under external torques $q_c, q_b$ were recorded, and $[q_c-q_a, q_b-q_a]$ was the range of backlash at target position $q_d$. We repeated the experiments at different joint positions and then used the mean values as the backlash joint range.

Except for the dynamics and backlash, another feature that needs to be model is the self-lock of the proximal finger joints. Since the proximal joints use a worm gear mechanism as motion transmission, the back driving is not allowed, which means they cannot move against the actuating direction. We simulated this feature by adjusting the corresponding joint positions and velocities.

\begin{figure}[t]
	\centering
	\subfloat{
    \includegraphics[trim=200 120 200 100, clip, width=0.30\linewidth]{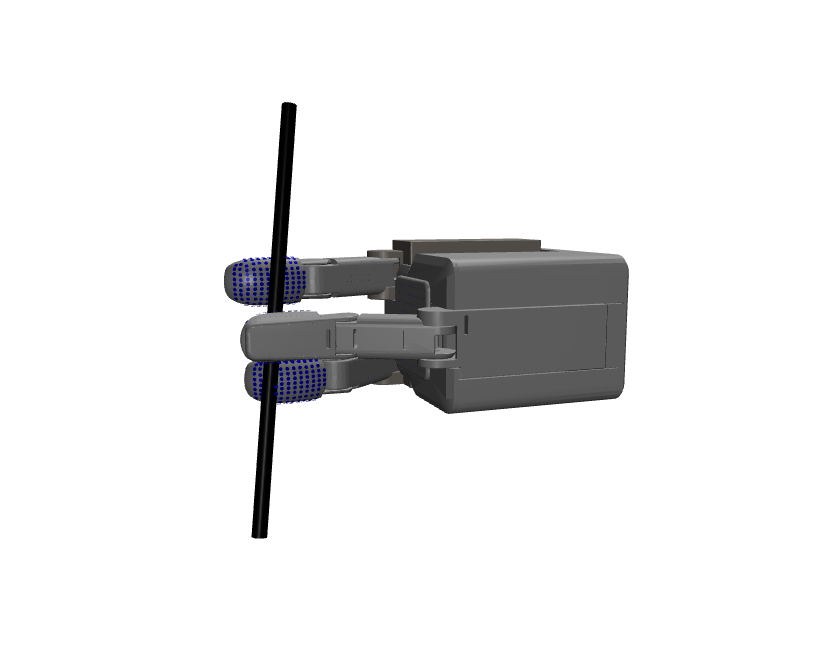}
	}
	\subfloat{
	\includegraphics[trim=200 120 200 100, clip, width=0.30\linewidth]{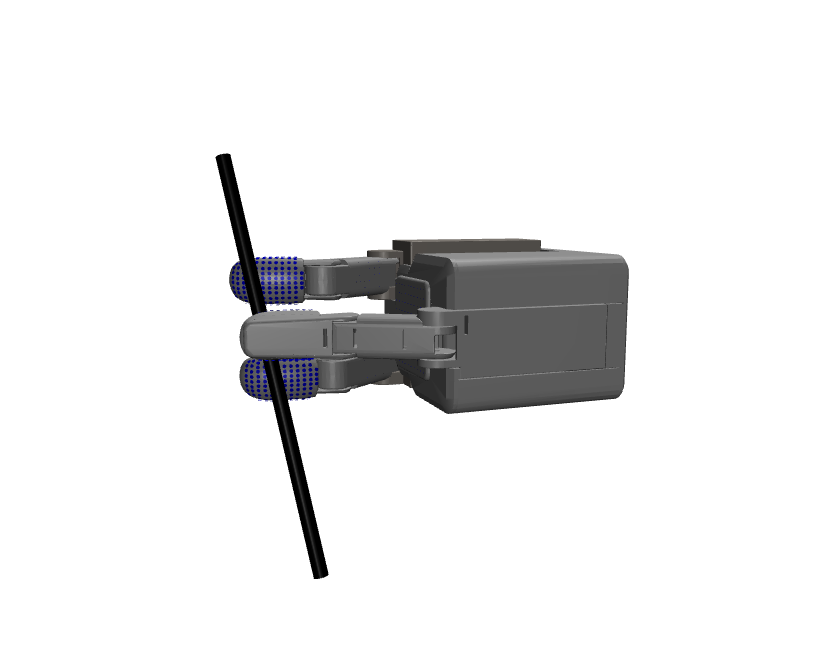}
	}
	\subfloat{
	\includegraphics[trim=200 120 200 100, clip, width=0.30\linewidth]{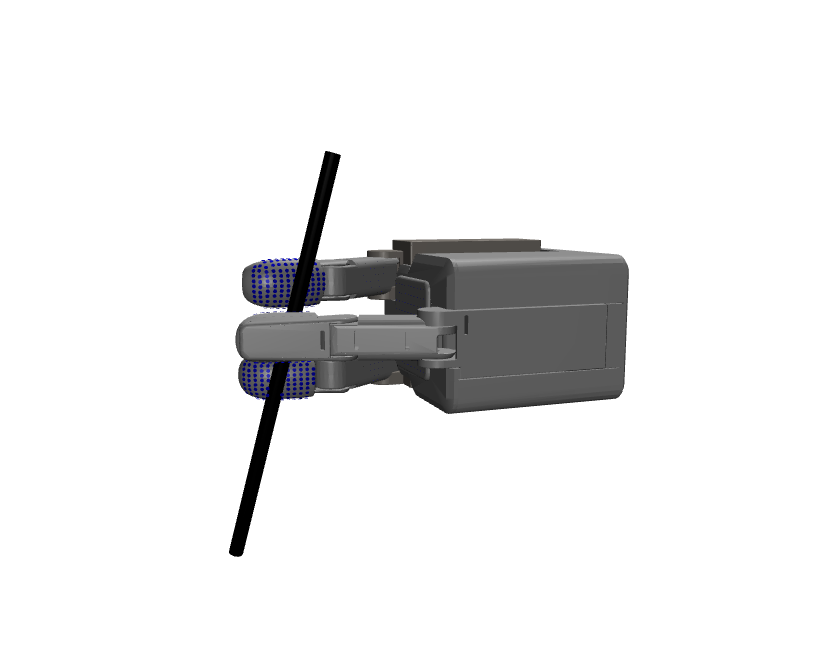}
	}	
	\caption{Examples of the generated initial states for the policy training. Fingers are in contact with the stick of randomized poses.}
	\label{fig:initial_states}
	\vspace{-4mm}
\end{figure}

\subsection{Tactile Sim-to-real}
Since we used the Boolean values to express whether the tactile sensors were in contact, we binarized the tactile signals with a threshold. Sensors that return values greater than the threshold were regarded as in contact, while the values below the threshold were filtered as noises.

Unlike the rigid-body simulation, the real tactile sensors are wrapped in deformable finger pulps. When the finger is in contact, the sensors \textit{around} but not \textit{in} the actual contact regions often return positive signals due to pressures from deformation. 
We used force-torque sensors to measure the sensitivities of tactile sensors and set the threshold of the tactile readings with an empirical value, which filtered out such signals and sensory noises but passed any signal from the sensors in contact. 


\subsection{Domain Randomization}
Besides the mean values of optimized dynamics parameters, the CMA-ES algorithm also provides their standard deviations. At the beginning of each training episode, we sampled the parameters, namely the PD gains of actuated joints and passive stiffness/damping of the backlash joints, from the optimized normal distributions. We also randomized other domain parameters to increase policy generalization ability, namely the stick weight, radius and the friction co-efficient between it and the fingers.

When training the dexterous manipulation skills with reinforcement learning, a common challenge is insufficient exploration of the state space, because the object is likely to be dropped at the early stage of training, and as the policy converges, the exploration becomes more limited.
Policies trained with deficient data space tend to be sensitive to disturbances and uncertainties.
To tackle such problems, we generated a data-set of random finger joint positions and object poses, and the training environment was initialized by sampling from the data-set at the beginning of each episode.

For better policy convergence and maneuverability, in candidate initial configurations, all fingers are in contact with the stick. The formed grasp does not have to be stable, because firstly it is an instantaneous state during a dynamic motion, and secondly the policy should learn to recover from the unstable state and continue the manipulation task. The proposed initial state generation method is described in \cref{alg:initial_state}.
We first sampled the finger joint positions and stick pose from uniform distributions, where the distribution ranges ensure the stick is between the fingers. Then we fixed the stick at the sampled pose, which cannot be moved by any external force. We closed the fingers with a constant speed until they contacted with the stick, and the current joint positions and stick pose were stored as one candidate initial state for policy training, as demonstrated in \cref{fig:initial_states}.
\begin{algorithm}[t]
\caption{Generating initial states for policy training.}
\label{alg:initial_state}
\textbf{Input} $\rho_\text{jnt}, \rho_\text{obj}$ \newline
Joint positions distribution, stick pose distribution
 
\textbf{Output} $\{\bm{q}_k, \bm{p}_k\}, k\in\{1,...,N\}$ \newline
A set for finger joint positions and stick poses
\begin{algorithmic}[1]
\While{$k \leq N$}                    
    \State {Sample from distributions: $\bm{q}_s\sim\rho_\text{jnt}, \bm{p}_s\sim\rho_\text{obj}$}
    \State {Set starting finger joint positions $\bm{q}_s$}
    \State {Fix the stick at the sampled pose $\bm{p}_s$}
    \State {Close the finger joints until contacts detected}
    \State {Add current data pair $\{\bm{q}_t,\bm{p}_t\}$ into the state set}
\EndWhile
\end{algorithmic}
\end{algorithm}

\begin{figure*}[t]
	\centering
	\subfloat{
    \includegraphics[trim=20 20 30 20, clip, width=0.23\linewidth]{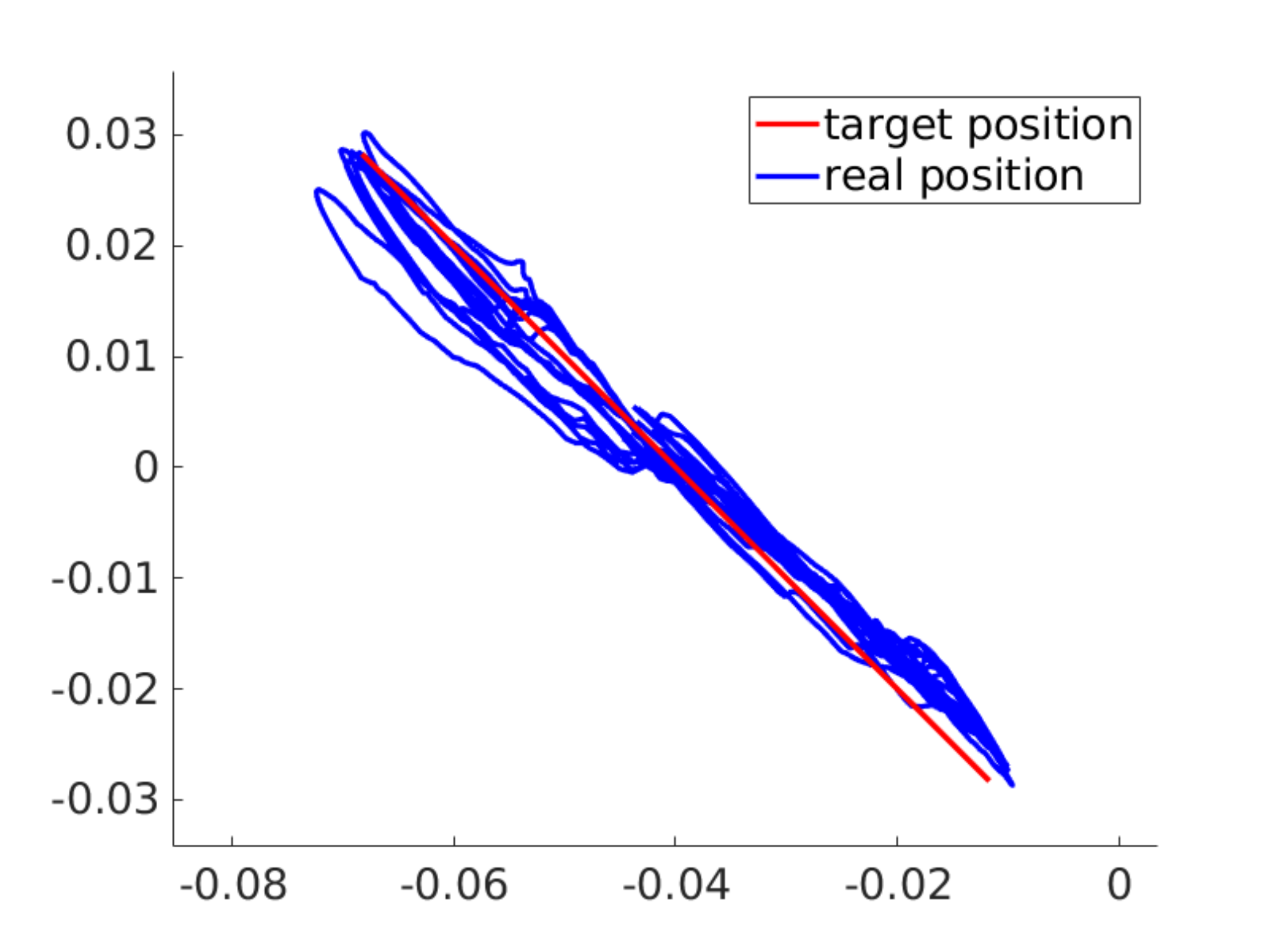}
	}
	\subfloat{
	\includegraphics[trim=20 20 30 20, clip, width=0.23\linewidth]{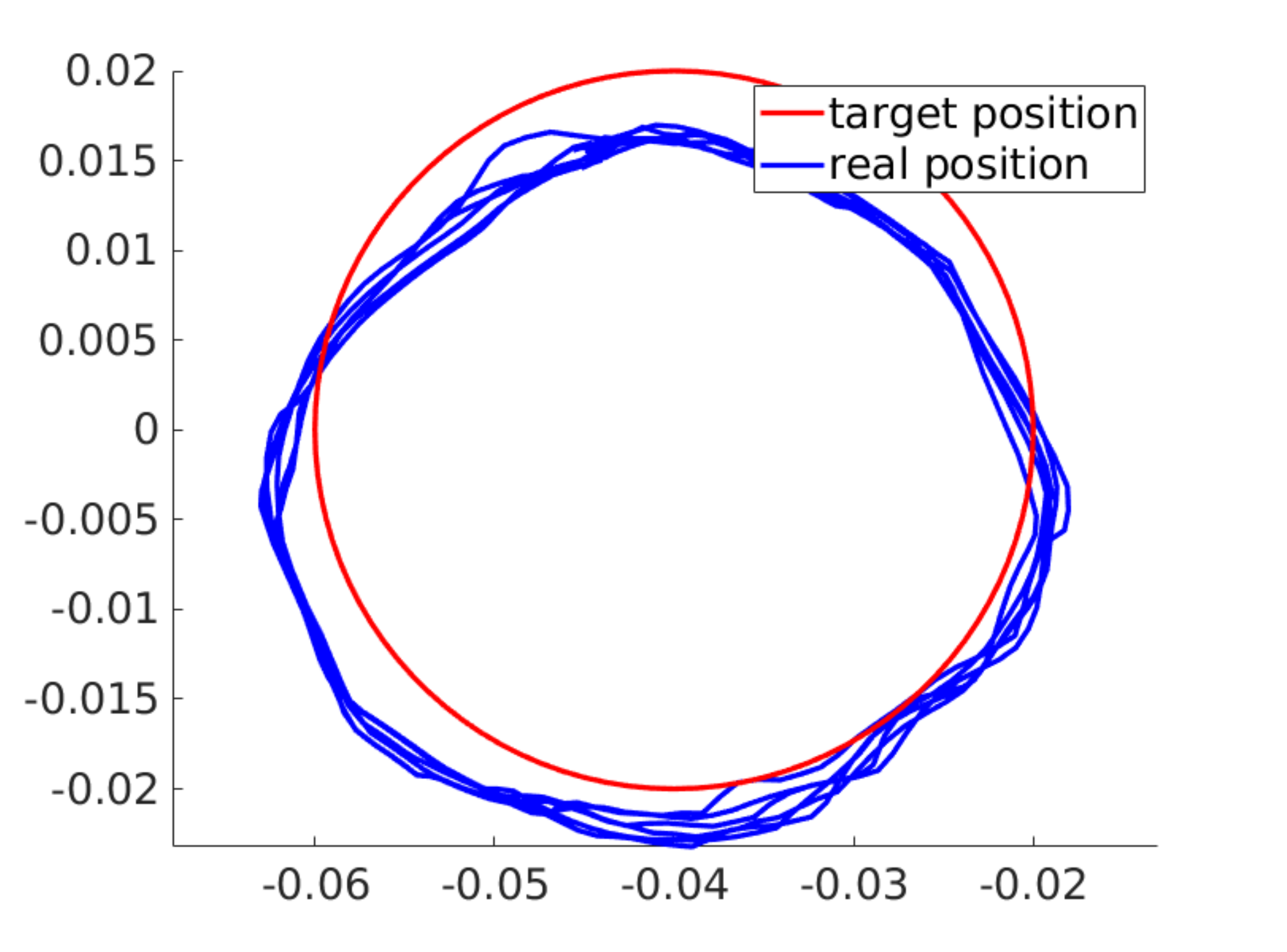}
	}
	\subfloat{
	\includegraphics[trim=20 20 30 20, clip, width=0.23\linewidth]{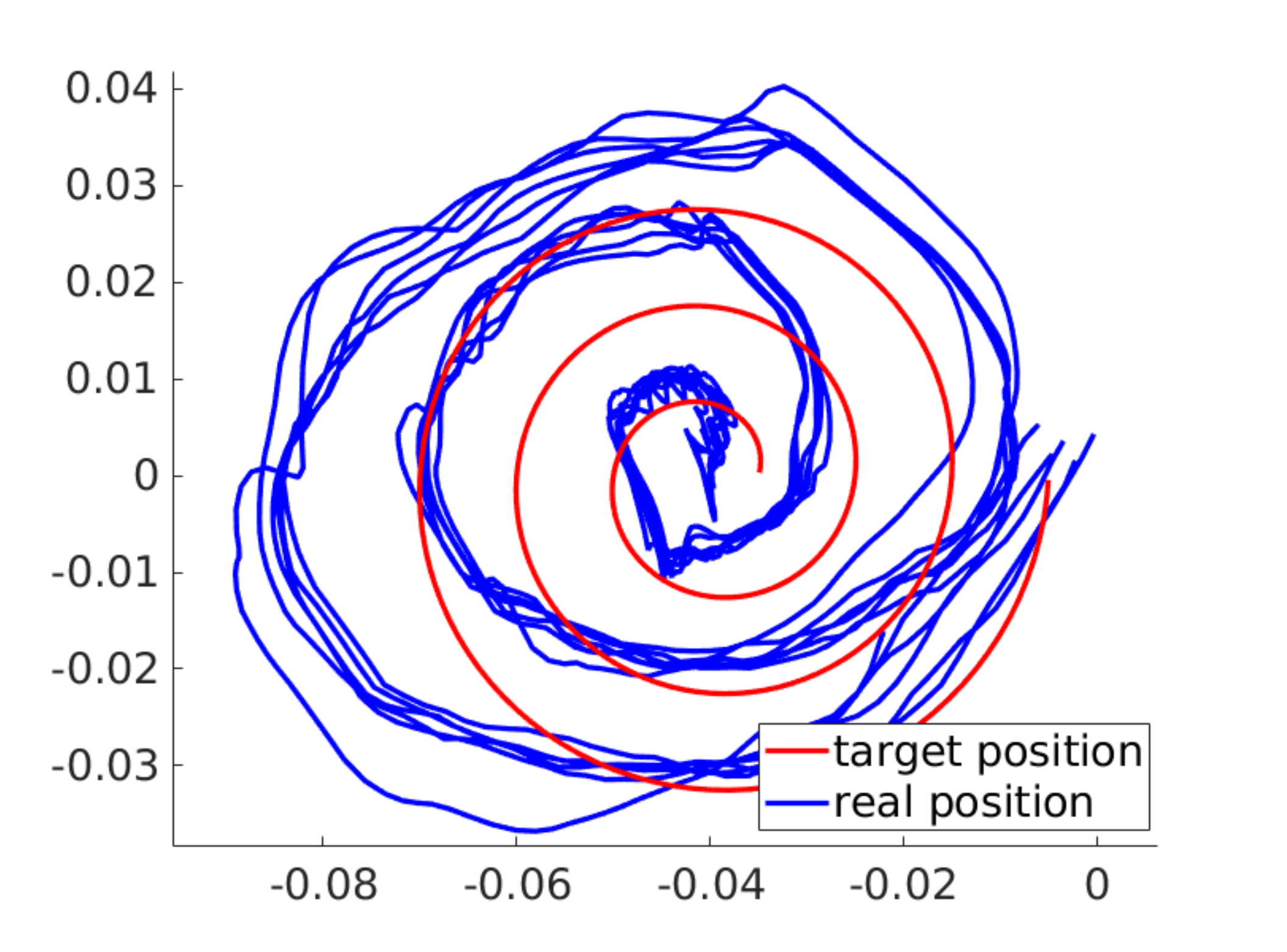}
	}
	\subfloat{
	\includegraphics[trim=20 20 30 20, clip, width=0.23\linewidth]{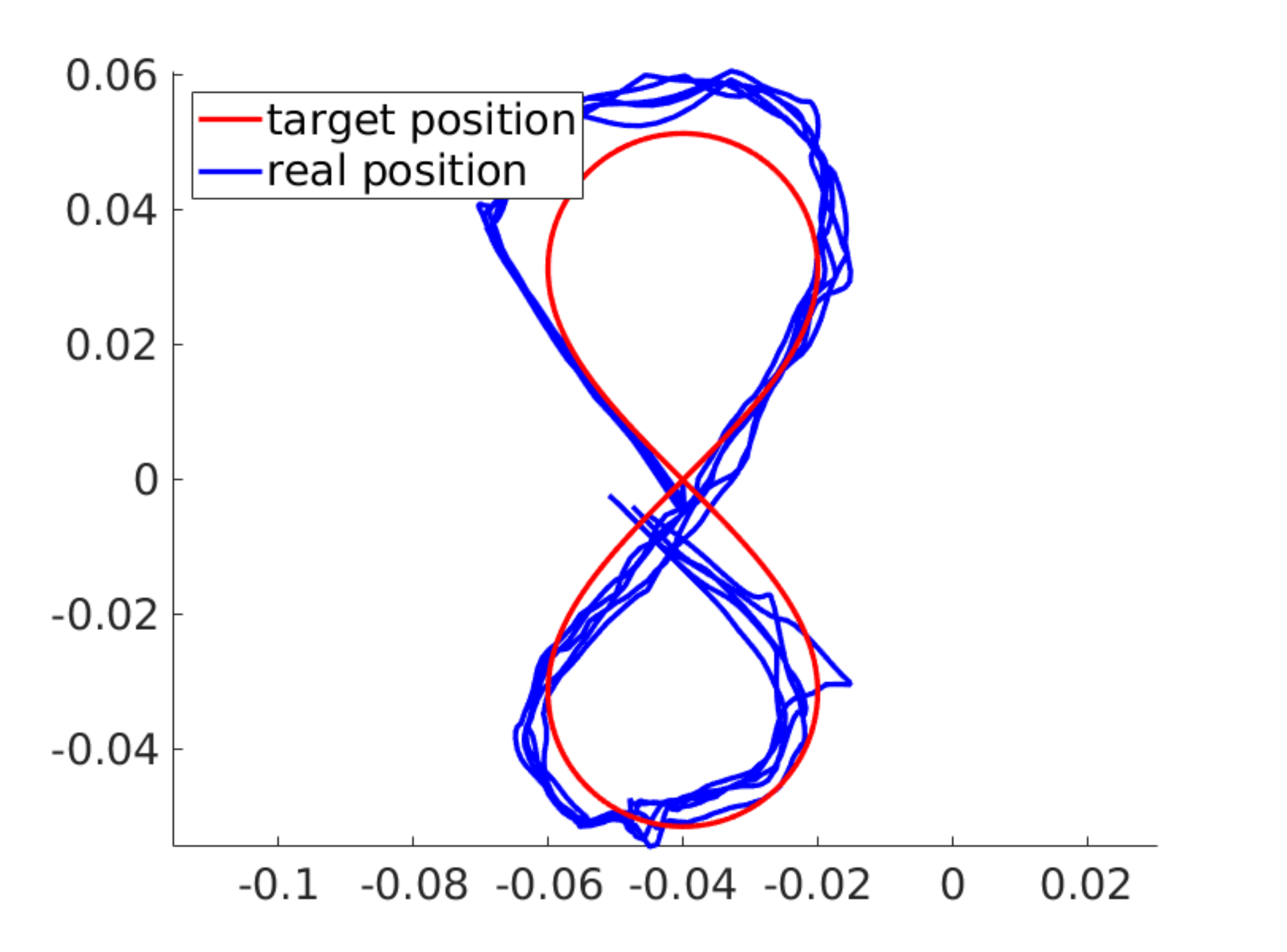}
	}
	\caption{The reference (red) and actual (blue) trajectories of the stick lower end-point on the X-Y plane. Four types of reference trajectories with increasing difficulties: the straight line, the circle, the spiral and the number eight. Each plot includes 10 episodes and each episode starts from the center. In the `circle' subplot we choose the second lap for clearness.}
	\label{fig:simulation_trajectories}
	\vspace{-4mm}
\end{figure*}

\section{Experiments} \label{sec:experiments}
In this section we presented the performances of learned policies in both simulated and real environments.
We validated the effectiveness of tactile feedback by comparing the policies trained with different observation on the object.

\subsection{Simulation Performance}

\begin{figure}[t]
    \centering
	\includegraphics[trim=30 7 0 7, clip, width=\linewidth]{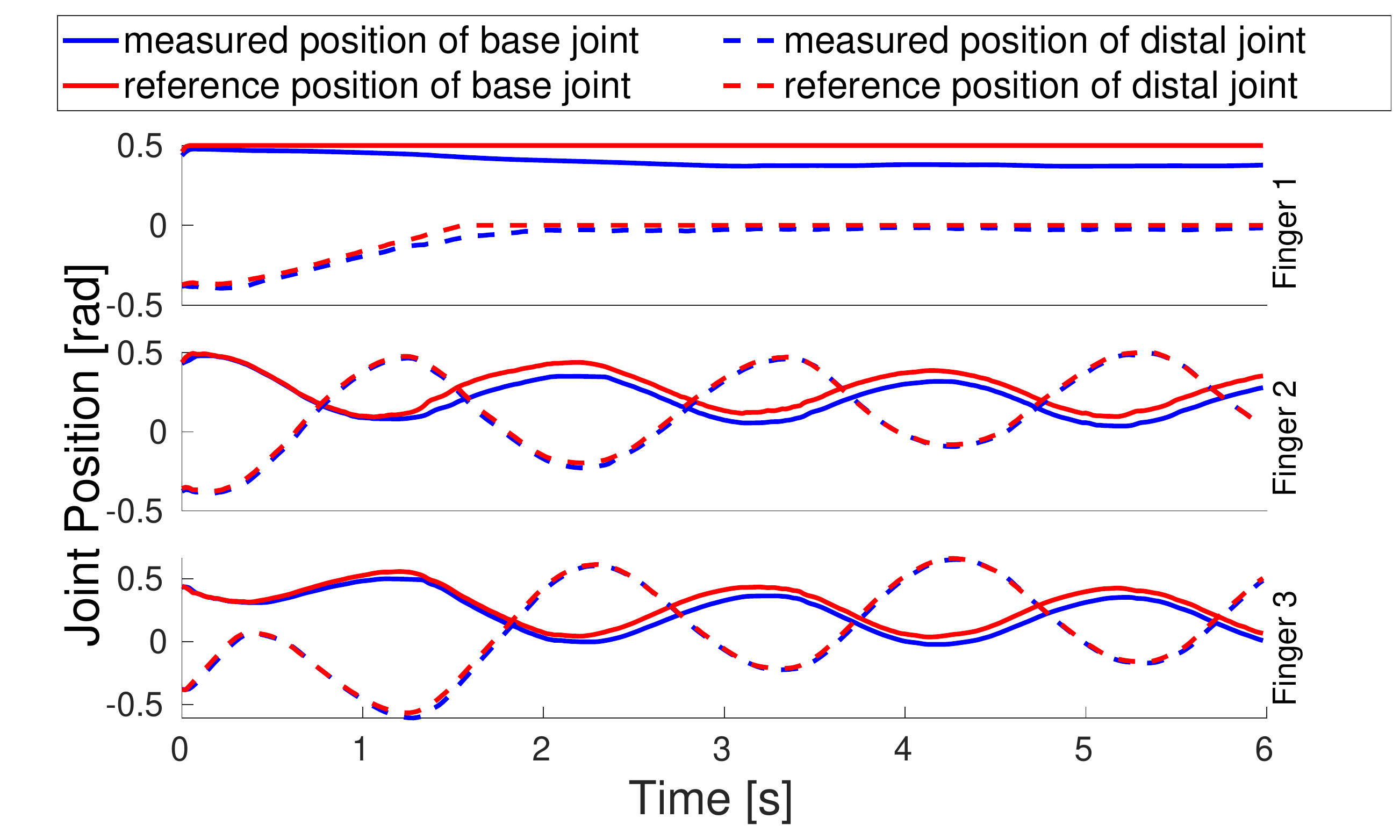}
	\vspace{-5mm}
    \caption{Reference and measured finger joint positions when manipulating the stick to draw a circle in simulation. 
 }
	\label{fig:joint_position}
	\vspace{-4mm}
\end{figure}

To validate the learning algorithm, we trained four policies with different reference stick pose trajectories, namely drawing the straight line, the circle, the number eight and the spiral with stick end-point. \cref{fig:simulation_trajectories} demonstrates reference and real trajectories of the stick lower end-point positions, showing that the trained policies are able to manipulate the stick to track the continuous pose references. The spiral trajectories are most challenging to the robot as it requires delicate maneuvers especially at the inner laps, thus the resultant trajectories have more errors.

To analyse the learned policies, we plotted the finger joint positions during the stick manipulation in simulation, as shown in \cref{fig:joint_position}. We chose the task of drawing the circle for its representativeness. The finger 1 (the thumb) almost remained still during the manipulation, providing a fixed supporting surface for the stick.
The finger 2 and 3 determined the motion of the stick, and the movements of their joints were periodical and highly correlated, demonstrating the coordination between different fingers and joints.

\cref{tab:errors} shows the average position errors of stick end-point and the orientation errors of the stick during different tasks.
Left part of the table displays the tracking errors under training settings, indicating that the learned policies using tactile feedback can manipulate the stick to track the reference pose in real-time.
The tracking errors increase as the ascent of the reference trajectory complexity, from drawing a straight line with random direction, to the trajectory of spiral.
To evaluate the generalization ability of the trained policies, we chose `circle' policy as an example.
In the training setting the desired angular velocity was $\pi/s$ and the desired radius was $2 cm$. We evaluated the policy with the settings of $[\frac{\pi}{2}/s, 2cm]$, $[\frac{3}{2}\pi/s, 2cm]$, $[\pi/s, 1cm]$, $[\pi/s, 3cm]$.
As shown in the right part of \cref{tab:errors}, the policy achieved comparable performances on circle trajectories with novel desired rotational velocities and radiuses.

\begin{table}[t]
	\centering
    \caption{Position and orientation errors of the stick in different manipulation tasks in simulation. Errors are averaged from 10 episodes each.}
    \label{tab:errors}	  
	\def\arraystretch{1.3}
	\begin{threeparttable}
	\begin{tabular}{>{\centering}p{1.5cm} | >{\centering}p{0.45cm} >{\centering}p{0.45cm} >{\centering}p{0.45cm}  >{\centering}p{0.45cm} | >{\centering}p{0.45cm} >{\centering}p{0.45cm} >{\centering}p{0.45cm} >{\centering}p{0.45cm}}
	    \hline
		& \multicolumn{4}{c}{Training Settings} & \multicolumn{4}{c}{Generalization Tests\tnote{*}} \tabularnewline
		\cline{2-9}
		& line & circle & spiral & eight & $v-$ & $v+$ & $r-$ & $r+$ \tabularnewline
		\hline
        $p_\text{err}$ [cm]  &0.83 &0.97 &2.32 &1.16 &1.19 &1.41 &0.85 &1.61\tabularnewline
        \hline
		$q_\text{err}$\tnote{\textdagger} \space [deg]  &1.80 &1.92 &5.90 &2.36 &2.23 &3.16 &1.70 &2.50\tabularnewline
		\hline 
	\end{tabular}
	\begin{tablenotes}
      \small
      \item[*] In the task of drawing a circle, we tested the trained policy with novel target rotational speed and radius.
      \item[\textdagger] Error on the orientation is the angle between desired and real direction vector of the stick.
    \end{tablenotes}
    \end{threeparttable}
	\vspace{-4mm}
\end{table}

\subsection{Sim-to-real Transfer}
To validate the trained policies in real world, we deployed them to the real robot hand with tactile sensors. The high level policy loop ran at 50 Hz, taking the measured joint positions and tactile information as feedback, and sending the commands to the low level joint position controller which ran at 1 kHz. Each trial started from a configuration where the stick was already grasped by the three fingertips as illustrated in \cref{fig:reference_pose}.
During the manipulation the robot hand was fixed at a certain pose by the robot arm. \cref{fig:screen_shot} demonstrates the snapshots of real robot in-hand manipulation experiments, showing that the policies trained with simulated data can be transferred to the real-world environments.
For the video of all experiments please refer to the supplementary materials.

Due to the deformation of materials wrapping the tactile sensors, the readings of some sensors could drift, returning positive values even if there is no contact. Since we binarized the tactile readings with a fixed threshold, such temporary sensor drifts could lead to errors in the discriminated contact states. 
Hence we calibrated the tactile sensors before every manipulation trial. We first logged the sensor readings without any contact in a period of time, and then used the mean values $\overline{s'}$ as the offsets. The calibrated tactile readings $s(t)$ were computed as:
\begin{equation}
    s(t) = \max(s^{*}(t) - \overline{s'}, \ 0)
\end{equation}
where $s^*$ denotes the raw tactile readings. We regarded the negative values as the invalid and set them to 0.

\begin{figure}[t]
    \centering
    \includegraphics[trim=100 10 110 0, clip, width=\linewidth]{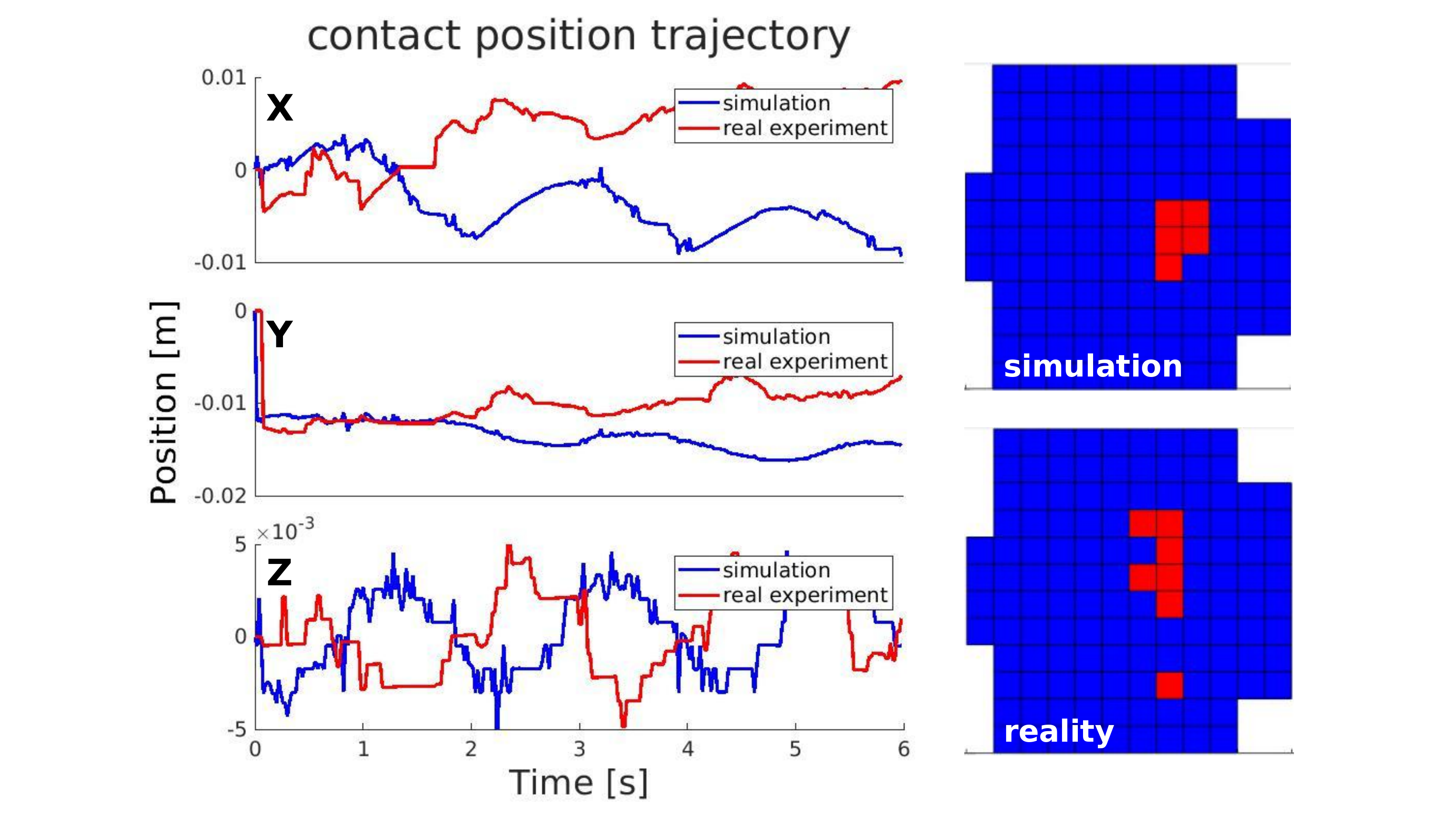}
    \vspace{-5mm}
    \caption{Left column: Contact position on a fingertip during the manipulation in both simulation (blue) and real-world experiment (red), which is averaged from the positions of activated tactile sensors, in fingertip local coordination (see \cref{fig:reference_pose}). Right column: Examples of activated tactile sensors (red blocks) in simulation and reality during the stick manipulation. The fingertip is pointing towards the right.
  }
    \label{fig:tactile_analysis}
    \vspace{-4mm}
\end{figure}

\begin{figure}
    \centering
    \includegraphics[trim=200 100 200 60, clip, width=0.75\linewidth]{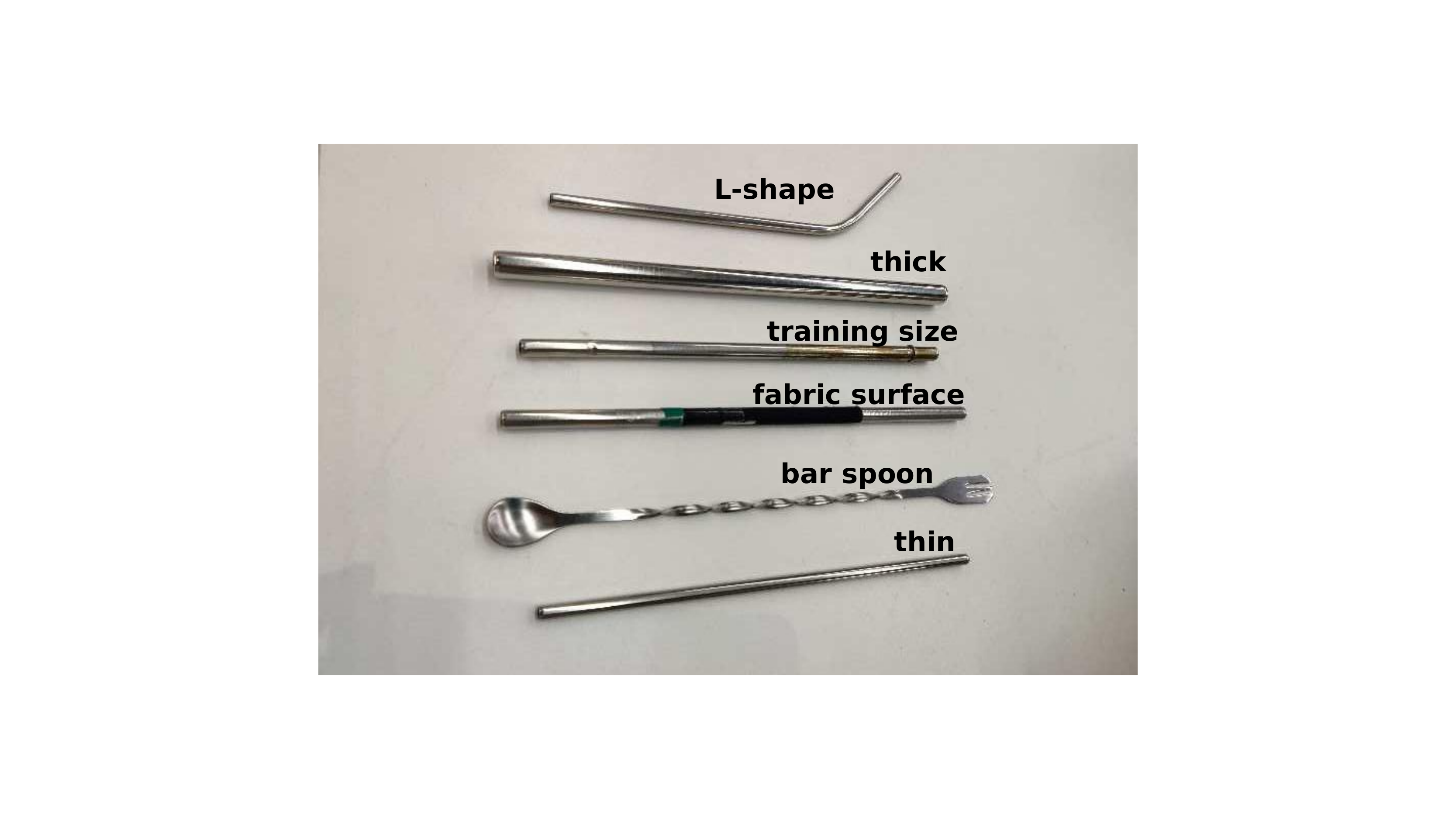} 
	\caption{Sticks with different shapes, weights and surfaces (stainless steel and fabric) are used to validate the policy in real-world experiments.}
	\label{fig:sticks}
	\vspace{-4mm}
\end{figure}

We compared the tactile feedback on one fingertip during the circle-drawing stick manipulation task in both simulation and reality, as shown in \cref{fig:tactile_analysis}.
Note that the contact position trajectories in simulation and reality are not expected to be consistent with each other because the initial stick poses were not the same.
Here we put the trajectories in the same plots just to conduct the qualitative analysis.
The contact positions were in local frame of the corresponding fingertip, as shown in \cref{fig:reference_pose}. Both real and simulated trajectories exhibited periodicities especially in the Z direction, indicating that the learned policies leveraged the curved finger surfaces to manipulate the stick, maintaining the contact position within the feasible range.
The right column of \cref{fig:tactile_analysis} illustrates the examples of activated tactile sensors. Compared with simulation, the detected contact region in real-world experiments had more irregular margins, due to the uncertainties in fingertip deformation and inconsistent sensitivities of different sensors.
However, the aforementioned differences in tactile feedback were instantaneous and the properties of real tactile sensors were consistent with simulation during most of the time.
Hence, though trained with pure simulated tactile information, the policies adapted to the tactile sim-to-real gap and performed well in real-world experiments.

We evaluated the generalization ability of the trained policies over different objects, as shown in \cref{fig:sticks}. Combined snapshots in the last row of \cref{fig:screen_shot} demonstrate the manipulation of the testing objects which are novel to the trained policies. With real-time tactile perception, though trained with a single stick in simulation, the policies can adapt to diversities in the stick shapes, weights and friction coefficients.

\subsection{Comparative Study}

\begin{figure}[t]
    \centering
    \includegraphics[trim=50 5 80 20, clip, width=\linewidth]{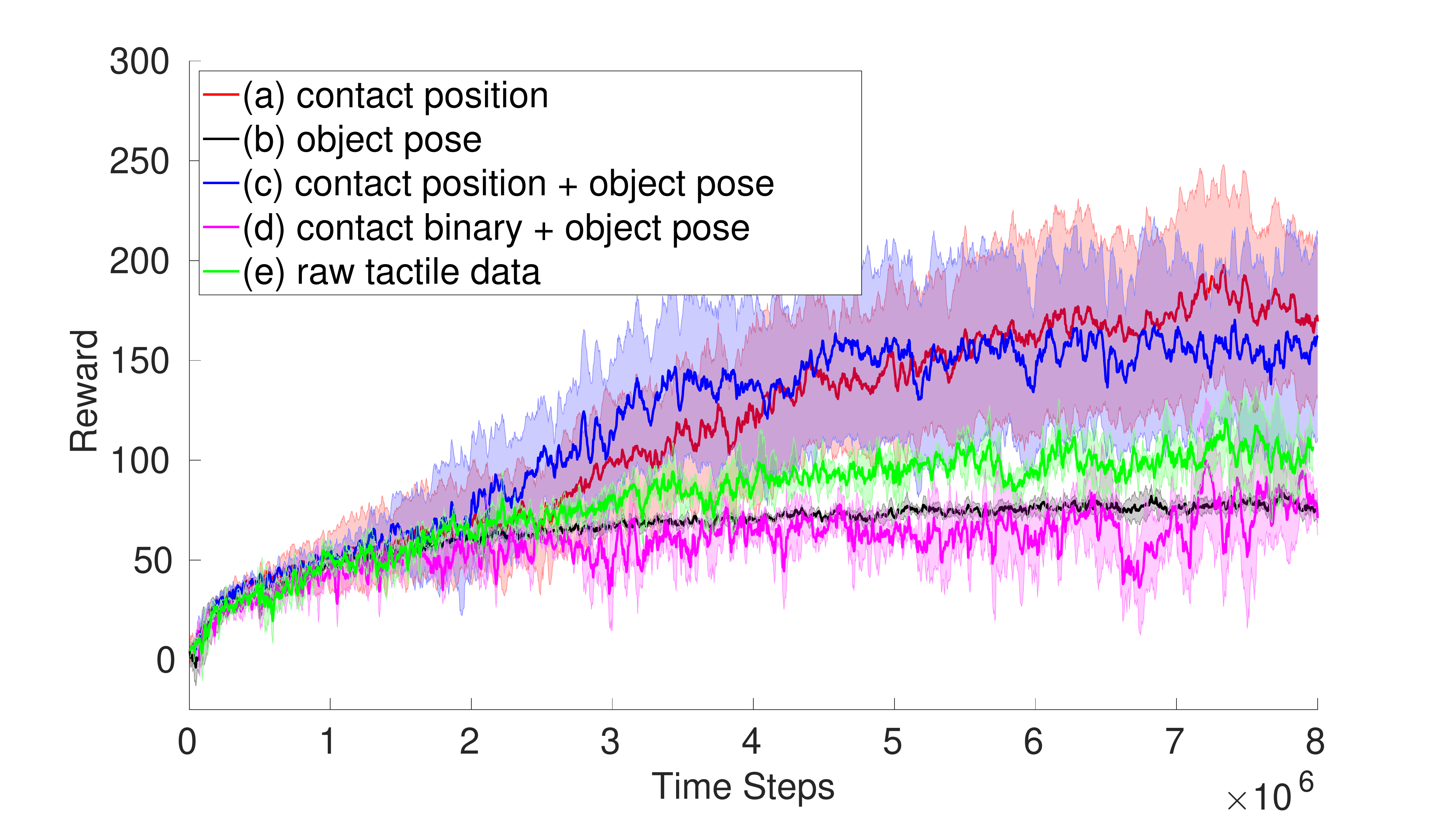}
    \vspace{-5mm}
    \caption{Learning curves of policies using different perceptions on the object. Training of each setting is repeated five times with random seeds. Policies using contact positions outperform others, and adding object pose does not further improve the performance.}
    \label{fig:comparison_learning_curve}
    \vspace{-4mm}
\end{figure}

To validate the effectiveness of the tactile information in learning dexterous in-hand manipulation, we conducted the comparative study where policies were trained with different observation spaces.
In comparison to (a) contact center positions on each finger, we trained the policies with (b) object pose, (c) object pose and contact center positions, (d) raw tactile information, (e) object pose and the binary information of whether each finger is in contact.
For each setting we trained the policy for three runs with random seeds and the average learning curve is shown in \cref{fig:comparison_learning_curve}.
Policies using contact center positions achieved the best performance, and adding the ground truth object pose into observation space did not provide further improvement. Moreover, policies trained with object pose solely could not achieve comparable performance, showing that the tactile information is more essential than visual information in in-hand manipulation tasks.

The tactile information of binary contact checks for three fingers ($1\times3$ dimensional) was uninformative for learning feasible manipulation policies, while the raw tactile data ($128\times3$ dimensional) required more complex network structure (e.g. graph convolutional networks) to distill the spatial information. Compared with them, the contact center positions ($3\times3$ dimensional) provided more straightforward and useful information, and achieved better tracking performance in stick rotation tasks with the same size of training data and network structure.

\begin{figure*}[t]
	\centering
    \includegraphics[trim=125 0 125 0, clip, width=\linewidth]{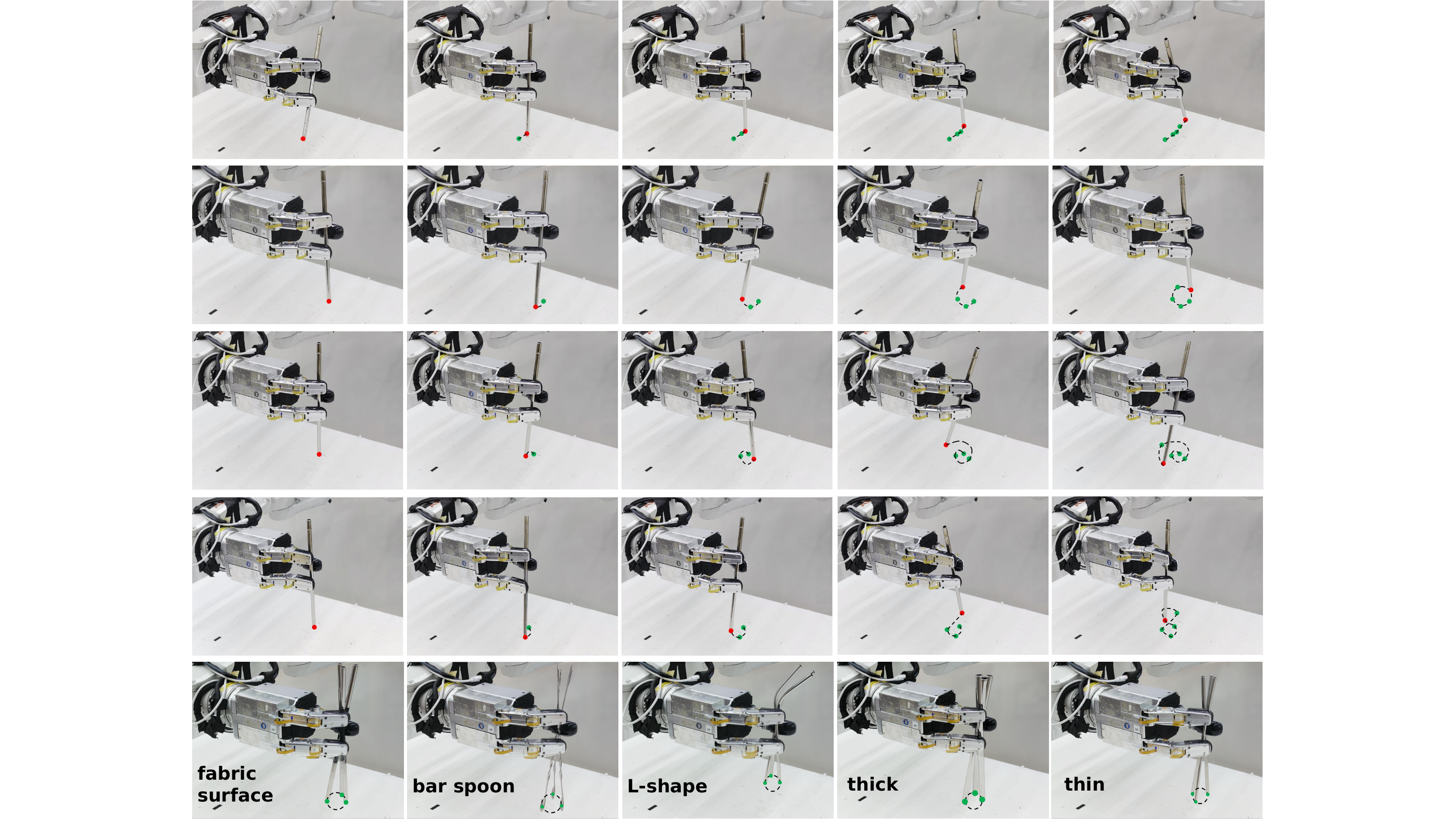}
	\caption{Snapshots of the real robot experiments. The first four rows represent the manipulation of tracking different trajectories, while the last row represents the manipulation of novel sticks shown in \cref{fig:sticks}. Note that the sticks were not contacted with the table. The red dots denote the current position of stick lower endpoint, while the green dots denote previous positions. The trajectories of the endpoint are illustrated as black dashed lines.}
	\label{fig:screen_shot}
	\vspace{-4mm}
\end{figure*}



\section{Conclusions and Future Work} \label{sec:discussion}

In this study, we achieved the tactile-based dexterous in-hand manipulation of slender cylindrical objects via deep reinforcement learning. Our approach estimates the contact center position on each fingertip, allowing the trained policies to track multiple real-time reference trajectories with only the point contracts of 3 robot fingers.

Due to the time-consuming and potential danger of gathering real robot training data, we used a physics simulator to train the policies and directly transferred them to the real robot. To reduce the sim-to-real gap, we calibrated the finger joints dynamics, binarized tactile signals, and randomized the initial training states. The trained policies performed well in both simulation and real-world experiments, demonstrating generalization to novel objects and tasks.

To evaluate the effectiveness of tactile feedback in in-hand manipulation, we conducted a comparative study, which trained policies with different object perception data. Policies trained with estimated contact center positions outperformed others, highlighting the importance of low-dimensional distilled tactile feedback for manipulating stick-like, slender cylindrical objects.

While in-hand manipulation of rigid objects allows the use of binarized tactile information, the performance can be enhanced further by using denser information. For example, continuous tactile readings provide not only contact regions but also the constant change of pressure distribution. Also, the ability to measure precisely the magnitude of normal forces of tactile sensors can indicate fingertip deformation and exerted forces during interactions. In future work, we plan to model the non-linear relation between tactile readings and surface pressure, and therefore exploit the use of continuous full tactile signals to achieve finer in-hand manipulation of a larger variety of objects.


\bibliographystyle{IEEEtran}
\balance
\bibliography{reference}

\vspace{-30pt}

\begin{IEEEbiography}[{\includegraphics[width=1in,height=1.25in,clip,keepaspectratio]{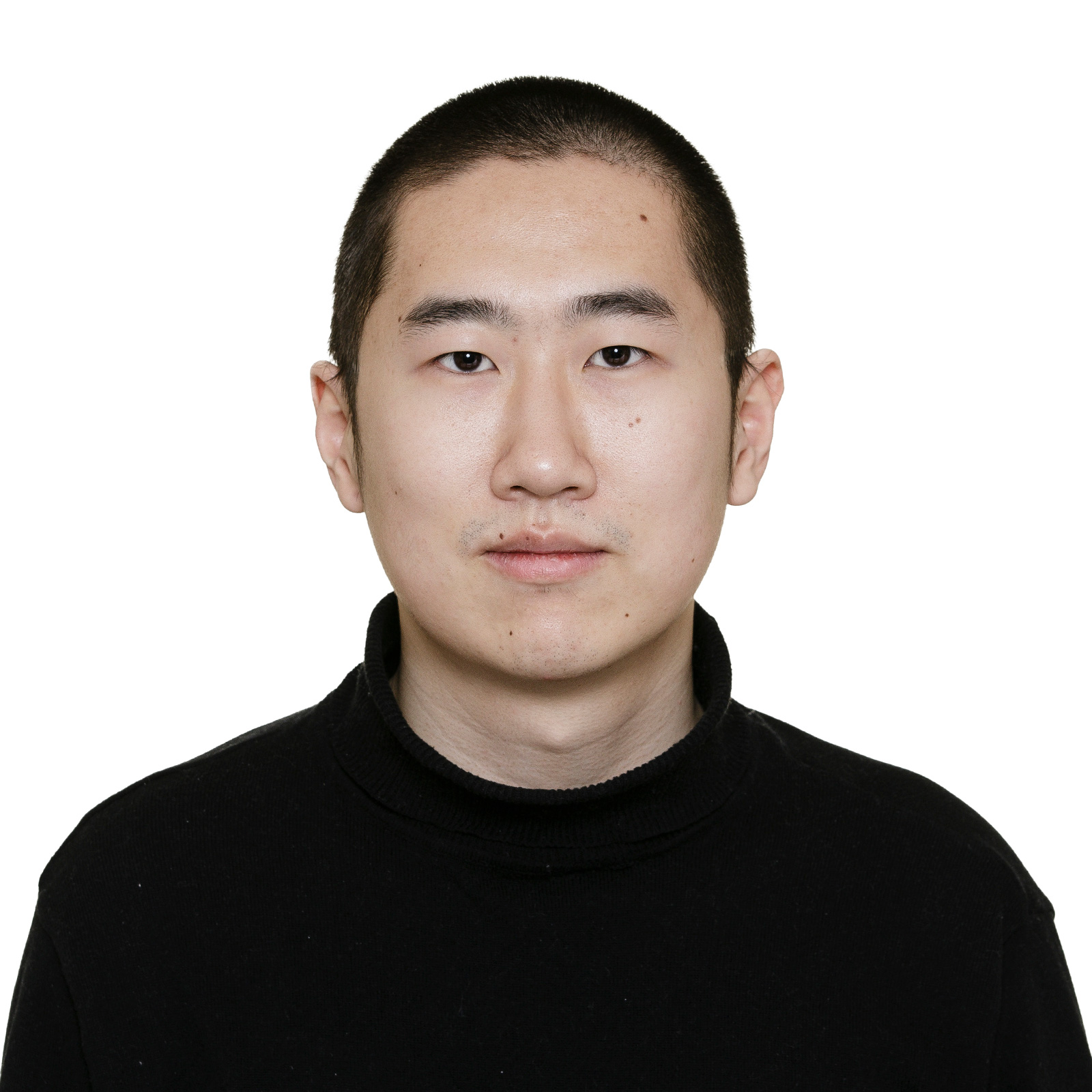}}]
{Wenbin Hu} received the B.S. degree in Automation from Tsinghua University, Beijing, China, in 2018. He is currently a Ph.D. student in robotics with the Advanced Intelligent Robotics (AIR) Lab, University of Edinburgh, UK.
His research interests include learning based planning and control of robotic grasping and manipulation.
\end{IEEEbiography}

\vspace{-30pt}

\begin{IEEEbiography}
[{\includegraphics[width=1in,height=1.25in,clip,keepaspectratio]{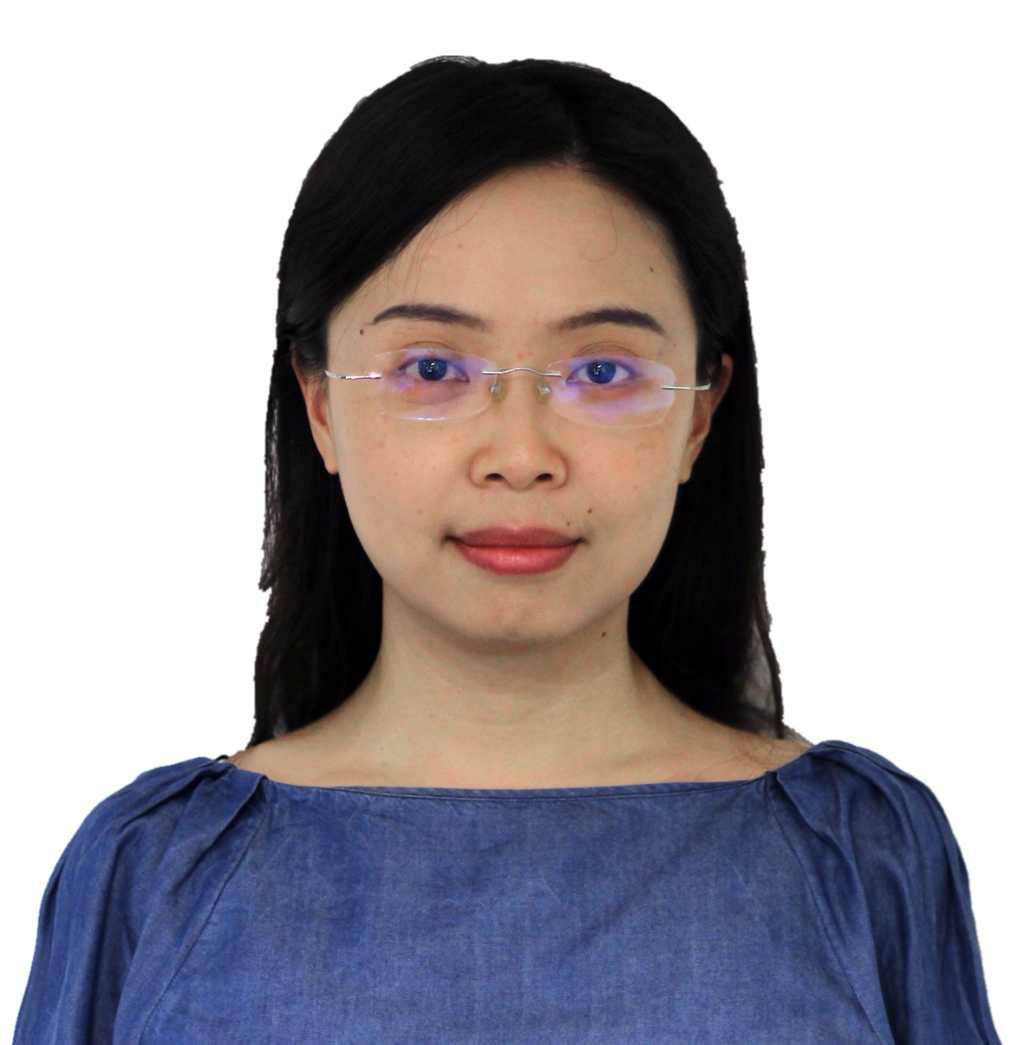}}]
{Bidan Huang} received her PhD degree in the University of Bath for her study on robotics in 2015. She is currently a senior researcher at Tencent Robotics X Lab. In 2015-2018, she worked as a research associate at the Hamlyn Centre for Robotic Surgery, Imperial College London. In 2012-2014, she was a visiting student of the Learning Algorithms and Systems Laboratory (LASA), Swiss Federal Institute of Technology in Lausanne (EPFL). Her main research interests are in robot learning, control, grasping and manipulation. 
\end{IEEEbiography}

\vspace{-30pt}

\begin{IEEEbiography}
[{\includegraphics[width=1in,height=1.25in,clip,keepaspectratio]{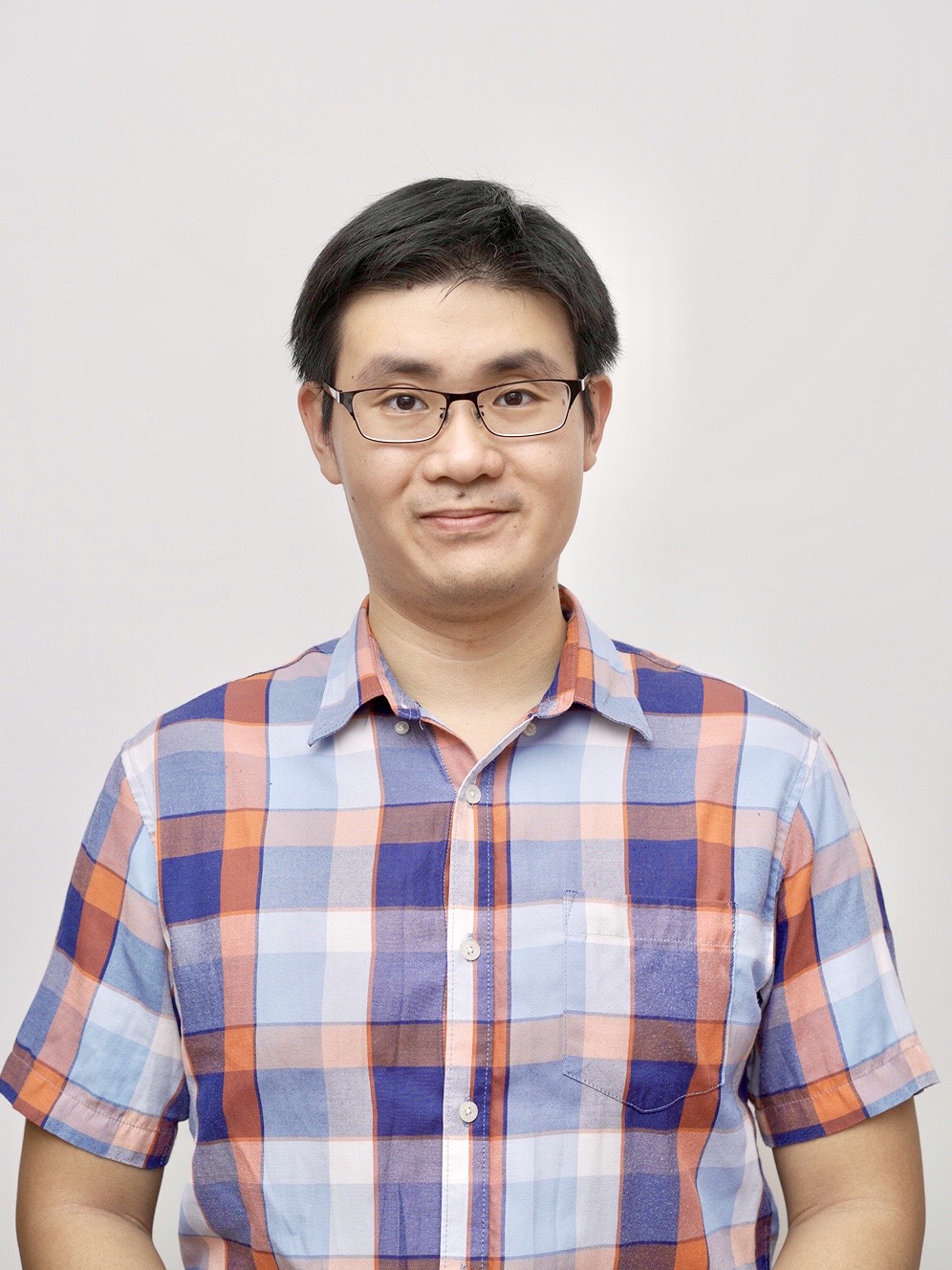}}]
{Wang Wei Lee} received his Ph.D. degree from the National University of Singapore (NUS) in 2016. He was a Postdoctoral Research Fellow with the Biomedical Institute for Global Health Research and Technology at NUS between 2016 and 2019. He joined Tencent Robotics X in April 2019 as an Advanced Research Scientist in charge of tactile sensor research. His research interests include the development of robust tactile sensing technologies applicable to real-world applications. 
\end{IEEEbiography}

\vspace{-30pt}

\begin{IEEEbiography}
[{\includegraphics[width=1in,height=1.25in,clip,keepaspectratio]{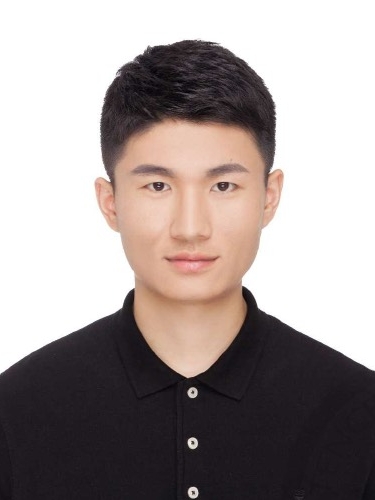}}]
{Sicheng Yang} received his B.E. degree in mechanical and automation engineering from Beihang University, China, in 2014 and the M.S. degree in mechanical engineering from Tsinghua University, China, in 2017. He joined Tencent Robotics X in October 2018 as an Advanced Research Scientist in charge of mechatronic system research. His research interests include dexterous hand-arm system, robotic grasping and manipulation and legged robot.
\end{IEEEbiography}

\begin{IEEEbiography}
[{\includegraphics[width=1in,height=1.25in,clip,keepaspectratio]{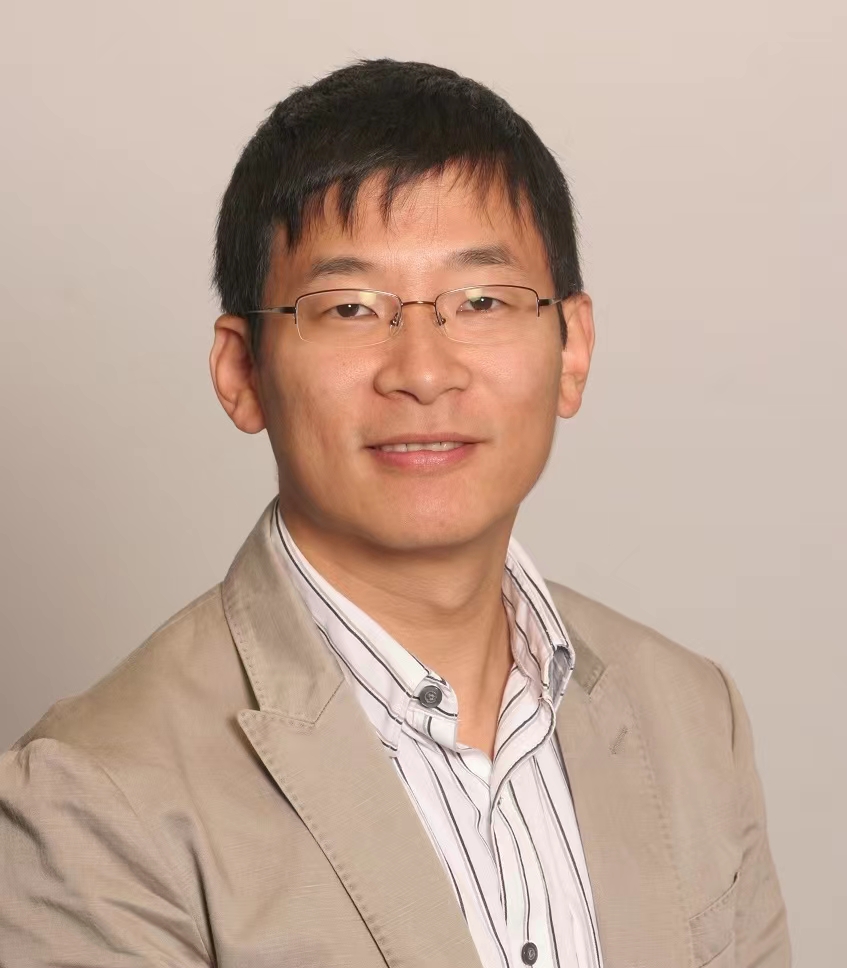}}]
{Yu Zheng} received the Ph.D. degree in mechatronics from Shanghai Jiao Tong University in 2007 and the Ph.D. degree in computer science from the University of North Carolina at Chapel Hill in 2014. 
From 2014 to 2018, he was an Assistant Professor with the Department of Electrical and Computer Engineering, University of Michigan-Dearborn. He joined Tencent Robotics X in 2018 as a Principal Research Scientist and Team Lead of the Control Center. His research interests include multi-body robotic systems, dexterous grasping and manipulation and legged locomotion. He serves as an Associate Editor for IEEE Robotics and Automation Letters.
\end{IEEEbiography}


\begin{IEEEbiography}[{\includegraphics[width=1in,height=1.25in,clip,keepaspectratio]{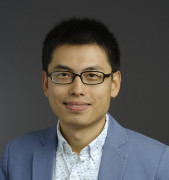}}]
{Zhibin(Alex) Li} received the joint Ph.D. degree in robotics from the Italian Institute of Technology, Genova, Italy and the University of Genova, Genova, Italy, in 2012. 
He is currently an Associate Professor with the Department of Computer Science, University College London, London, U.K. His research interests include inventing new control, optimization, and deep learning technologies, and creating intelligent behaviors of dynamical systems with human comparable abilities to move and manipulate.
\end{IEEEbiography}


\end{document}